\PassOptionsToPackage{dvipsnames}{xcolor}
\documentclass[11pt,letterpaper]{style}
\usepackage[numbers]{natbib}
\usepackage{graphicx}
\usepackage{booktabs}
\usepackage{amsmath,amsfonts,amssymb}
\usepackage{cleveref}
\usepackage{subcaption}
\usepackage{wrapfig}
\usepackage{multirow}
\usepackage{colortbl}
\usepackage{listings}
\usepackage{xparse}
\usepackage{fontawesome5}
\usepackage{bxcoloremoji}
\usepackage{float}
\usepackage{placeins}
\usepackage{threeparttable}

\graphicspath{{./}{fig/}{figures/}{plot/}{pdf/}{table/}}
\usepackage{amsthm}
\usepackage{tcolorbox}
\usepackage{svg}
\tcbuselibrary{skins,breakable}
\tcbuselibrary{listingsutf8}
\usepackage{titletoc}

\usepackage{setspace}
\usepackage{pifont}
\usepackage{mathtools}
\usepackage{enumitem}
\usepackage{arydshln}
\usepackage{bbm}
\usepackage{lineno}
\usepackage{makecell}
\usepackage{adjustbox}
\usepackage[ruled,vlined]{algorithm2e}
\SetKwInput{KwRequire}{Require}
\SetKwInput{KwOutput}{Output}

\definecolor{mygray}{gray}{0.9}
\definecolor{syncol}{RGB}{243,246,249}
\definecolor{wildcol}{RGB}{215,240,235}
\definecolor{drop1}{RGB}{180,225,220}
\definecolor{drop2}{RGB}{150,210,200}
\definecolor{drop3}{RGB}{120,195,185}
\definecolor{drop4}{RGB}{95,180,170}
\definecolor{drop5}{RGB}{65,160,150}
\definecolor{lightblue}{RGB}{210,230,250}

\newcommand{\up}[1]{\textcolor{green!60!black}{{\small$\uparrow$#1}}}

\definecolor{myblue1}{HTML}{0171DC}
\definecolor{myblue2}{HTML}{013978} 

\NewDocumentEnvironment{minted}{O{} m +b}{%
}{}

\newcommand{\equal}{\textsuperscript{*}}     
\newcommand{\advisor}{\textsuperscript{\dag}}

\renewcommand\Authfont{\centering\normalfont\bfseries\fontsize{11}{15}\selectfont}
\renewcommand\Affilfont{\centering\normalfont\fontsize{10}{15}\selectfont}

\newcommand{\afflogo}[2]{%
  \raisebox{#2}{\includegraphics[height=1.5em]{#1}}%
}

\title{Reasoning Within the Mind: Dynamic Multimodal Interleaving in Latent Space}
\runningtitle{Reasoning Within the Mind: Dynamic Multimodal Interleaving in Latent Space}

\author{%
    {\Authfont
    \textbf{Chengzhi Liu}\equal\textsuperscript{1} \quad
    \textbf{Yuzhe Yang}\equal\textsuperscript{1} \\ 
    \textbf{Yue Fan}\textsuperscript{3} \quad
    \textbf{Qingyue Wei}\textsuperscript{2} \quad
    \textbf{Sheng Liu}\advisor\textsuperscript{2} \quad
    \textbf{Xin Eric Wang}\advisor\textsuperscript{1}
    }\\
    {\Affilfont
    \textsuperscript{1} \afflogo{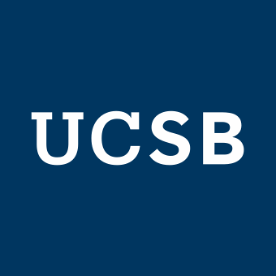}{-0.7ex} University of California, Santa Barbara \quad
    \textsuperscript{2} \afflogo{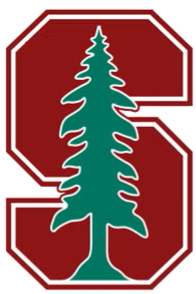}{-0.7ex} Stanford University \\
    \textsuperscript{3} \afflogo{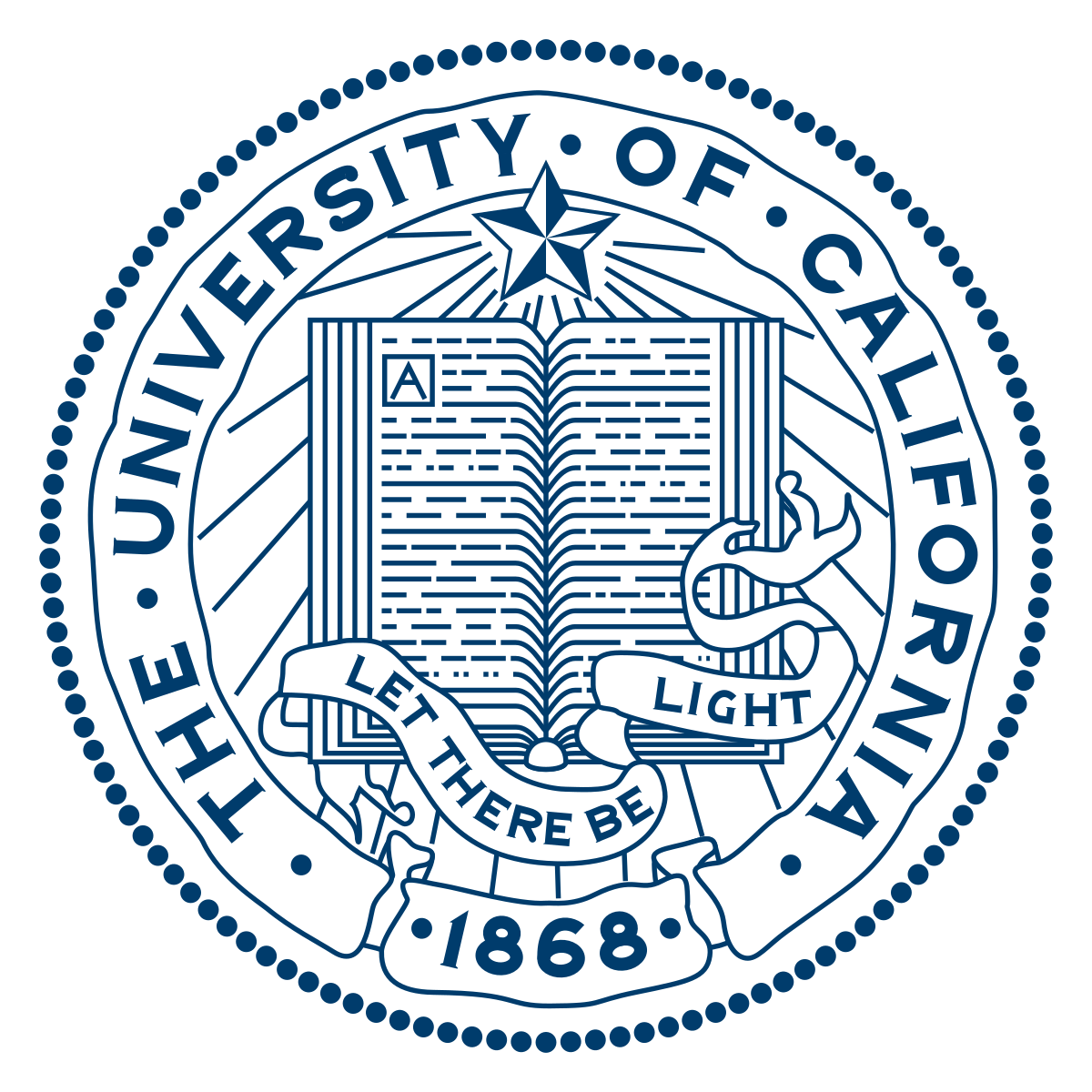}{-0.7ex} University of California, Santa Cruz \\
    \equal\ Equal contribution \quad \advisor\ Equal advising
    \par
    \texttt{\{chengzhi,yuzheyang,ericxwang\}@ucsb.edu, shengl@stanford.edu}\\
    \vspace{8pt}
    }
}

\begin{document}
\begin{abstract}
\textbf{Abstract:} Recent advancements in Multimodal Large Language Models (MLLMs) have significantly enhanced cross-modal understanding and reasoning by incorporating Chain-of-Thought (CoT) reasoning in the semantic space. Building upon this, recent studies extend the CoT mechanism to the visual modality, enabling models to integrate visual information during reasoning through external tools or explicit image generation. However, these methods remain dependent on explicit step-by-step reasoning, unstable perception–reasoning interaction and notable computational overhead. Inspired by human cognition, we posit that thinking unfolds not linearly but through the dynamic interleaving of reasoning and perception within the mind. Motivated by this perspective, we propose \textbf{{\color{myblue2}DMLR}}, a test-time \textbf{{\color{myblue2}\underline{D}ynamic \underline{M}ultimodal \underline{L}atent \underline{R}easoning}} framework that employs confidence-guided latent policy gradient optimization to refine latent think tokens for in-depth reasoning. Furthermore, a Dynamic Visual Injection Strategy is introduced, which retrieves the most relevant visual features at each latent think token and updates the set of best visual patches. The updated patches are then injected into latent think token to achieve dynamic visual–textual interleaving. Experiments across seven multimodal reasoning benchmarks and various model architectures demonstrate that DMLR significantly improves reasoning and perception performance while maintaining high inference efficiency.

\end{abstract}
\newcommand{\TitleLinks}{%
\centering
    \vspace{8pt}
    {\noindent\absfont\fontsize{11}{13}\selectfont
    \faGithub\ Project Page: \url{https://mllm-dmlr.github.io/}\par}%
}
\maketitle

\begin{figure*}[h]
\centering
\includegraphics[width=1\linewidth]{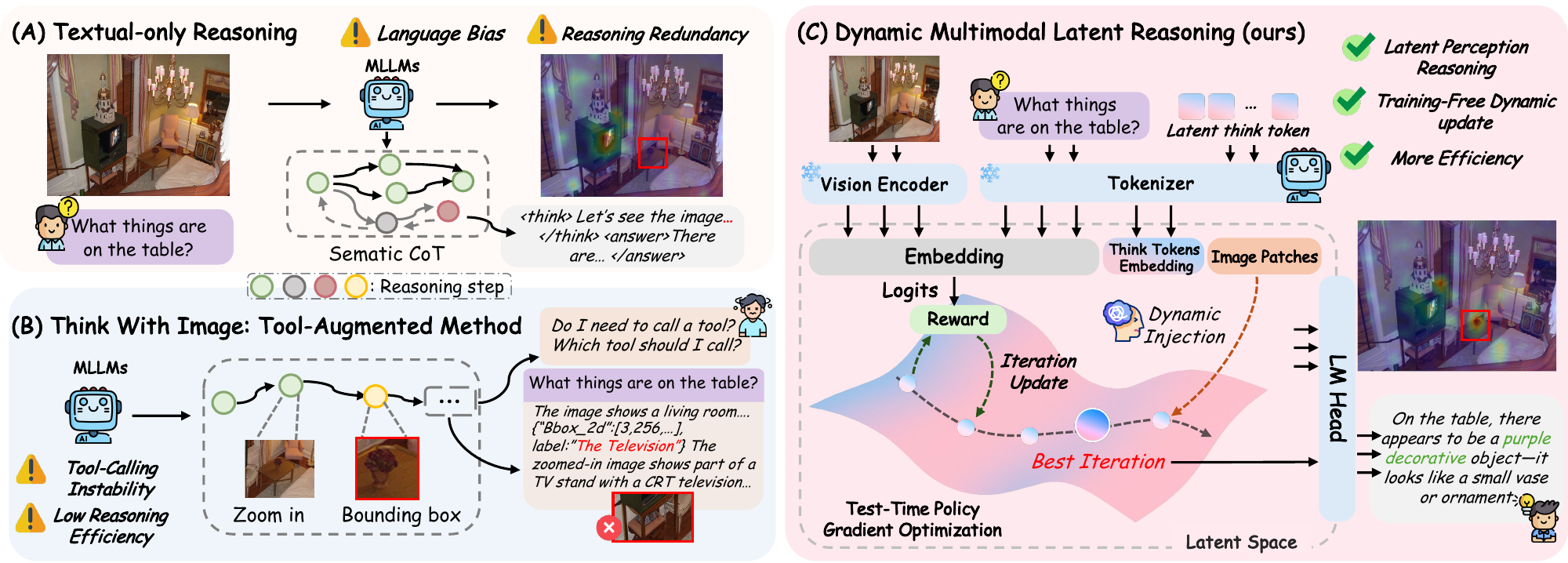}
\caption{\small Comparison between DMLR and two reasoning paradigms. (A) Text-only reasoning: relies solely on explicit CoT, often causing visual grounding errors and redundant steps. (B) Think-with-Image reasoning: depends on external perception tools, leading to unstable tool calls and extra overhead. (C) DMLR (ours): refines latent think tokens in the latent space through confidence-guided optimization and dynamically injects visual information, achieving self-improving reasoning without additional training while maintaining high efficiency.}
\vspace{-0.7cm}
\label{fig:intro}
\end{figure*}

\section{Introduction}
\label{sec:intro}

Multimodal Large Language Models (MLLMs)~\cite{bai2025qwen2, wang2025internvl3, vteam2025glm45vglm41vthinkingversatilemultimodal, li2024llava} have achieved remarkable breakthroughs in integrating visual and linguistic information. This progress has facilitated the incorporation of Chain-of-Thought (CoT) reasoning into multimodal tasks, enabling models to construct structured reasoning paths across visual and textual modalities. Current multimodal reasoning approaches can be broadly categorized into three types:
 \textit{(1) Textual-only Reasoning}~\cite{mondal2024kam, su2025thinking, huang2025vision}, which generates intermediate reasoning steps in the sematic space. Such methods explicitly express reasoning logic through language generation but often suffer from language bias and insufficient visual grounding, as shown in Figure.~\ref{fig:intro}(a).
\textit{(2) Think with Images} attempts to directly manipulate or augment images during reasoning, such as local zooming~\cite{su2025pixel, zheng2025deepeyes}, region highlighting~\cite{fu2025refocus, fan2025gritteachingmllmsthink}, or generating intermediate reasoning steps via diffusion models~\cite{li2025imagine, zhang2025latentsketchpadsketchingvisual} to enhance visual alignment. Despite their effectiveness in improving reasoning to a certain extent, they still face challenges such as unstable tool invocation and high inference overhead, as reflected in Figure~\ref{fig:intro}(b). Recently, latent-space reasoning has emerged as a promising paradigm for enhancing reasoning capabilities in large language models, as exemplified by approaches such as CoCoNut~\cite{hao2025traininglargelanguagemodels} and LatentSeek~\cite{li2025seekdarkreasoningtesttime}.
 Its core idea is to perform implicit reasoning in the latent space, replacing explicit textual steps with latent vectors to reduce redundant generation and capture more compact information. However, recent studies~\cite{li2025latentvisualreasoning,yang2025machinementalimageryempower, pham2025multimodalchaincontinuousthought, zhang2025memgenweavinggenerativelatent}  still rely on extra training to enforce latent reasoning triggered at fixed positions (via special tokens). This rigidity prevents the model from adaptively allocating reasoning effort.

Inspired by human cognition, we argue that reasoning is not fixed. Instead, humans dynamically revisit visual information, specifically when they encounter uncertainty. Drawing on this intuition, we empirically analyze the interplay between the model's visual reliance and its internal confidence. Our analysis reveals two key phenomena: \textit{\textbf{(i)} Visual information is used only at a few specific stages of the reasoning process rather than at fixed positions}, and \textit{\textbf{(ii)} Internal confidence serves as a natural indicator for the need of visual grounding as it strongly correlates with reasoning correctness}. These findings suggest that effective multimodal reasoning relies on dynamic visual usage guided by internal confidence.


In light of these observations, we propose \textbf{DMLR}, a Test-time \textbf{D}ynamic \textbf{M}ultimodal \textbf{L}atent \textbf{R}easoning Framework, as shown in Figure~\ref{fig:intro}(c). Specifically, it introduces optimizable latent think tokens to serve as a mental draft, which are iteratively refined through confidence-guided policy gradient updates. Crucially, we design a confidence-driven dynamic visual injection strategy. At each step, the model autonomously determines whether to revisit visual information and which contents to select (ranging from none to a few specific patches). This mechanism allows the model to naturally skip visual injection when internal confidence is sufficient, or actively integrate targeted visual clues when necessary, all driven by the objective of maximizing reasoning confidence, effectively mimicking the human cognitive process of checking visual clues to build confidence. After several iterations, the optimized latent tokens are decoded with the input without extra inference cost. Extensive experiments demonstrate that DMLR consistently outperforms existing methods across diverse architectures and tasks while maintaining high efficiency. The main contributions can be summarized as follows:

\begin{itemize}[leftmargin=*]
\vspace{-1em}
    \item[\ding{182}] We reveal two key phenomena: Visual information contributes only at specific reasoning steps; and confidence reflects both reasoning quality and visual grounding.
    \item[\ding{183}] We propose DMLR, a test-time framework for multimodal latent reasoning that integrates confidence-guided latent optimization with dynamic visual injection.
    \item[\ding{184}] Extensive evaluations show that DMLR consistently outperforms other methods across diverse architectures and multimodal tasks, while maintaining high efficiency.
\end{itemize}

\section{Related Work}
\textbf{Explicit Reasoning.} Many prior works have explored visual reasoning. Early approaches mainly relied on semantic CoT, where the model performs all inference in the text space after a one-time visual encoding~\cite{mondal2024kam,li2024llava, chen2025sft, wang2025vlrethinkerincentivizingselfreflectionvisionlanguage}. However, this separation between perception and reasoning often leads to misalignment and hallucination~\cite{su2025thinking,liu2025thinkingseeingassessingamplified,liu2025reducing,tong2024eyeswideshutexploring, liu2025thinkingseeingassessingamplified, Tang_2025_CVPR}. To address these limitations, recent studies adopt a Thinking-with-Images paradigm, where the model can draw auxiliary elements~\cite{zhang2025latentsketchpadsketchingvisual, su2025openthinkimglearningthinkimages, hu2024visualsketchpadsketchingvisual}, zoom or crop regions~\cite{fan2025gritteachingmllmsthink,su2025pixel, hong2025deepeyesv2agenticmultimodalmodel,zhang2025chain}, or generate intermediate visual cues~\cite{zou2025looktwiceanswermemoryspace,fu2025refocus,zheng2025deepeyes}, enabling it to reason directly over visual structures.

\noindent \textbf{Latent Reasoning.} Recently, an increasing number of studies have begun to shift reasoning from the explicit token space to the model's latent representation space. Some methods introduce dedicated training frameworks that optimize latent representations to support more effective internal reasoning~\cite{liu2025fractional, hao2025traininglargelanguagemodels, huang2025thinkactvisionlanguageactionreasoningreinforced,mi2025milrimprovingmultimodalimage,deng2025latentreasoningllmsvocabularyspace, huang2025spotlighttokenperceptionmultimodal}, while others propose training-free approaches that manipulate latent activations during inference to refine the reasoning process~\cite{ li2025seekdarkreasoningtesttime,zhang2025softthinkingunlockingreasoning,butt2025softtokenshardtruths,ye2025thinkingflytesttimereasoning,li-etal-2025-feature}. In addition, several recent works explore injecting visual information into the latent space~\cite{li2025latentvisualreasoning, yang2025machinementalimageryempower,sun2025latentchainofthoughtvisualreasoning,gao2025interleavedmodalchainofthought,pham2025multimodalchaincontinuousthought}, enabling models to iteratively operate over both latent semantic features and latent visual cues, thereby supporting a more flexible form of interleaved multimodal reasoning.
\section{Preliminary and Motivation}
\label{motivation}
As shown in Figure~\ref{fig1}, existing reasoning paradigms commonly suffer from insufficient visual grounding, unstable tool invocation, and high computational overhead.
These limitations motivate a fundamental question: why can’t MLLMs reason like humans do, dynamically deciding how to reason and which visual information to pay attention on during the thinking process? To this end, we organize the section around two research questions:  \textit{\textbf{(RQ1)} Whether multimodal models require visual perception at every step of reasoning?}   \textit{\textbf{(RQ2)} If not, can their internal representations indicate when visual perception and reasoning is required?}

\subsection{\textbf{Dynamic Perception-Reasoning is Necessary} }

\textbf{Definition 3.1 \textit{(Visual Dependency Score)}.} 
\textit{Let the visual input be denoted as $I$, and its perturbed version as $\tilde{I}$.
Given a query $q$, the model’s dependence on visual information can be quantified by measuring the output discrepancy between the original and perturbed visual inputs.
Specifically, for the $i$-th generated sequence $\mathcal{X}_i = \{x_{i,1}, x_{i,2}, \dots, x_{i,t}\}$, the visual dependency score at position $t$ is defined as:}

\begin{equation}
S_{i,t} = \log \pi_\theta(x_{i,t}\mid x_{i,<t},I,q) - \log \pi_\theta(x_{i,t}\mid x_{i,<t},\tilde{I},q)
\end{equation}

\begin{wrapfigure}{r}{0.5\linewidth}
\vspace{-1.3em}
    \centering
    \includegraphics[width=\linewidth]{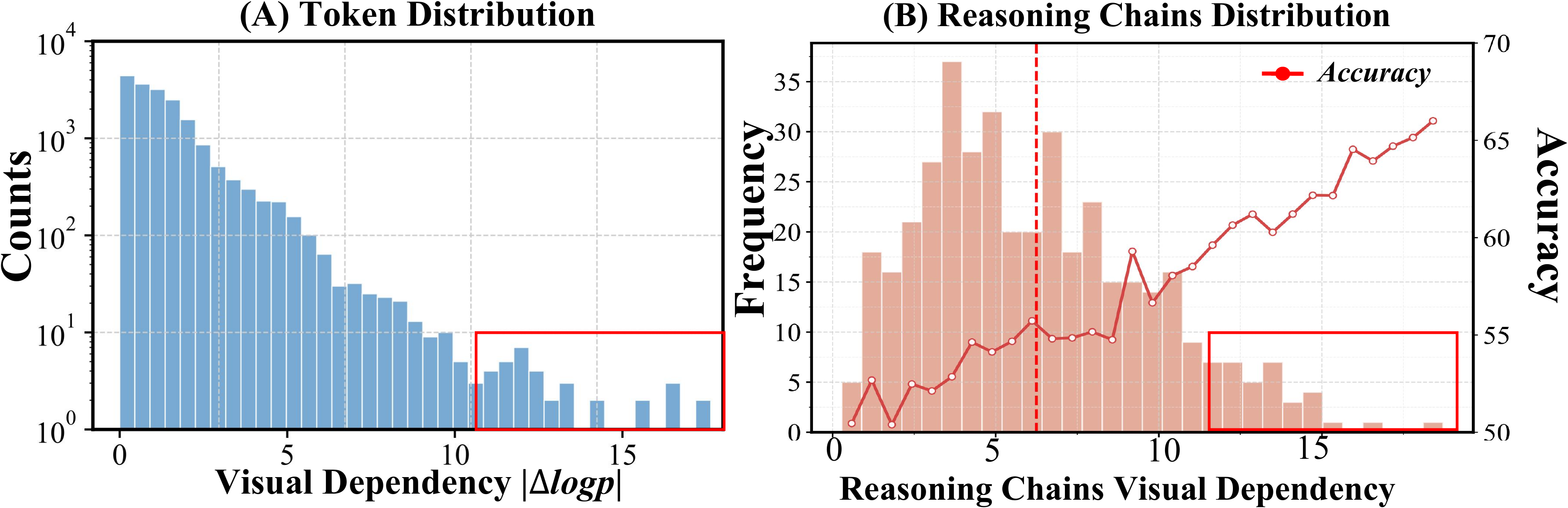}
    \caption{\small Analysis of visual dependency in reasoning. (A) Token-level distribution shows visual sensitivity is concentrated in a few tokens. (B) Chain-level distribution reveals large variation in visual reliance across reasoning trajectories.}
    \label{visual_dep}
\vspace{-2em}
\end{wrapfigure}

where $\pi_\theta(\cdot)$ denotes the token-level conditional probability distribution of the model. A larger $S_{i,t}$ indicates a stronger dependency of the generated token on visual information. Building upon the above metric, we analyze visual dependency on the MathVision benchmark using the Qwen2.5-VL-7B~\cite{bai2025qwen25vltechnicalreport} at two levels. First, for individual reasoning chains, we compute token-level visual dependency scores, capturing how much each generated token relies on visual information, as illustrated in Figure~\ref{visual_dep}(a). Second, as shown in Figure~\ref{visual_dep}(b), we aggregate these scores across full reasoning trajectories to obtain chain-level visual dependency, which reveals how different reasoning paths vary in their reliance on visual perception. These results reveal that:

\noindent \ding{70}~\textbf{\textit{Takeaway 1.}} \textit{The dependency on visual input across the reasoning process is highly uneven: only a small subset of tokens show strong sensitivity to visual features, while the majority operate independently of the image.}

\noindent\ding{70}~\textbf{\textit{Takeaway 2.}}  \textit{Across reasoning chains sampled from the same model, visual dependency varies substantially. Chains exhibiting stronger visual reliance consistently yield higher accuracy.}

\subsection{\textbf{Internal Confidence Affects Multimodal Reasoning}}
\label{con222}

\textbf{Definition 3.2 \textit{(Confidence Gain)}.} \textit{Let $I$ denote the visual input, $q$ the query, and
$\mathcal{T}_{t}$ denotes the reasoning at step $t$. The Confidence Gain at step $t$ is defined as the change in the probability of the ground-truth answer $Y_{gt}$ after adding step $x_t$. A positive $\mathcal{G}_t$ suggests that step $x_t$ strengthens the confidence, 
whereas a negative value indicates the opposite.}
\begin{equation}
\mathcal{G}_t = 
\log \pi_\theta (Y_{gt} \mid I, q, \mathcal{T}_{\le t})
- \log \pi_\theta (Y_{gt} \mid I, q, \mathcal{T}_{< t})
\label{eq:icr}
\end{equation}

\begin{figure}[h]
\centering
\begin{minipage}{0.48\linewidth}
    \centering
    \includegraphics[width=\linewidth]{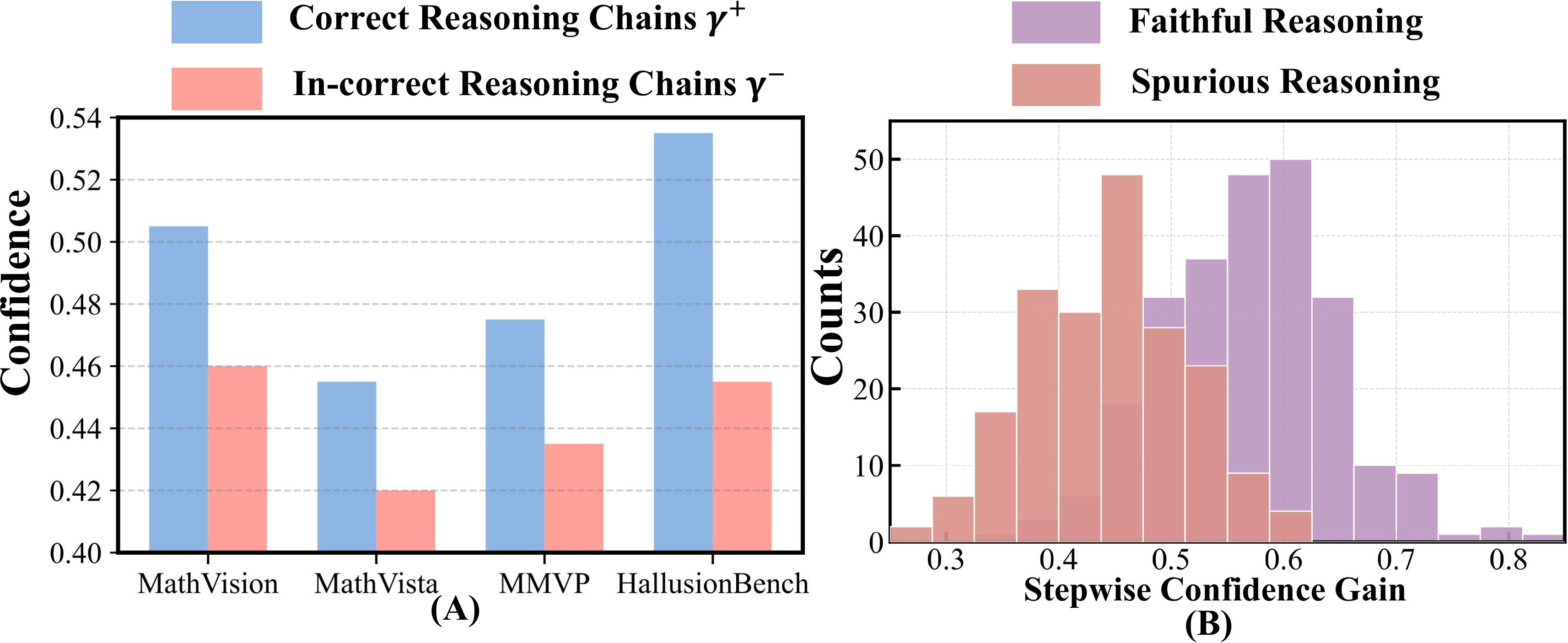}
    \captionof{figure}{\small Analysis of the relationship between confidence and reasoning quality. (A) Correct reasoning chains exhibit substantially higher frequencies of positive confidence gains than incorrect ones. (B) Faithful reasoning shows consistently stronger confidence improvement than spurious reasoning.}
    \label{confidence}
\end{minipage}
\hfill
\begin{minipage}{0.48\linewidth}
    \centering
    \includegraphics[width=\linewidth]{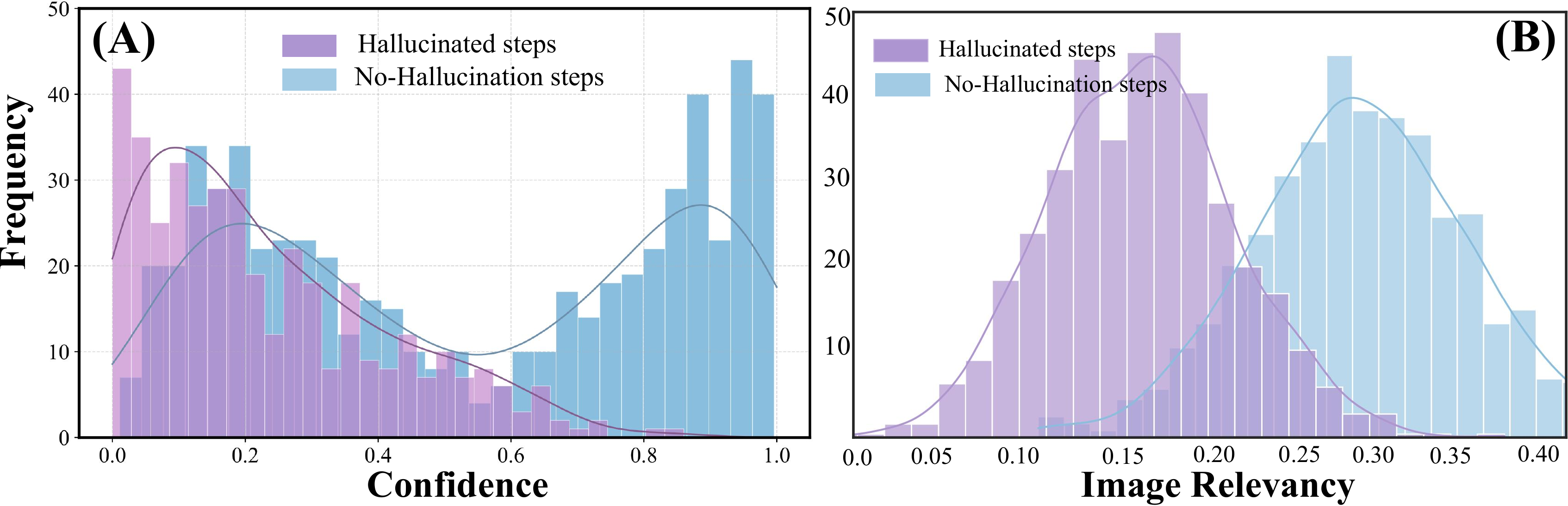}
    \captionof{figure}{\small Analysis of the relationship between confidence and visual grounding. (A) Hallucinated steps show lower confidence than non-hallucinated ones. (B) Hallucinated steps exhibit weaker image relevancy than their counterparts.}
    \label{halu}
\end{minipage}
\end{figure}

\noindent \ding{118}~\textbf{ \textit{Observation 1: Higher Confidence Tends to Indicate Higher Reasoning Accuracy.}} We analyze reasoning chains generated by various reasoning models across four benchmarks,  where all chains are partitioned into a correct set $\mathcal{T}^{+}$ and an incorrect set $\mathcal{T}^{-}$ based on their answer correctness. We then compute the proportion of reasoning steps for each chain that obtain a positive confidence reward. As shown in Figure~\ref{confidence}(a), reasoning chains in $\mathcal{T}^{+}$ exhibit a substantially higher proportion of positive confidence increments compared to those in $\mathcal{T}^{-}$,  \textit{indicating that the reasoning leading to correct answers tends to exhibit more stable and higher confidence.}

\noindent \ding{118}~\textbf{\textit{Observation 2:  Confidence Reflects Reasoning Chains Quality.}} We investigate whether confidence dynamics reflect reasoning quality by evaluating reasoning chains within the correct set $\mathcal{T}^{+}$ using the evaluator GPT-4o~\cite{openai2024gpt4ocard}. Each chain is assessed for logical validity and factual consistency, and categorized into Faithful and Spurious groups. As shown in Figure~\ref{confidence}(b), faithful reasoning chains exhibit a higher proportion of positive confidence increments, \textit{suggesting that  confidence improvement not only correlates with answer accuracy but also reveals the intrinsic quality of the reasoning process.}

\noindent \ding{118}~\textbf{\textit{Observation 3: High Confidence Aligns with Stronger Visual Grounding.}} We further evaluate various reasoning models on the perception benchmark to analyze the relationship between confidence and visual grounding. Each step in the reasoning chain is categorized as \emph{hallucinated} or \emph{non-hallucination} based on whether it refers to an object actually present in the image. As shown in Figure~\ref{halu}, hallucinated steps exhibit lower confidence and weaker visual grounding, while non-hallucinatory steps maintain higher and more stable confidence with stronger visual alignment. \textit{The results indicate that confidence acts as an intrinsic signal of visual faithfulness, with higher confidence consistently associated with more reliable reasoning.}

\section{Methodology}
\subsection{Problem Formulation}
Given a text input sequence $\mathcal{Q} = (q_1, \dots, q_k)$ and a set of visual embeddings $\mathcal{Z} = (z_1, \dots, z_I)$ extracted by a visual encoder, 
the MLLM $\pi_{\theta}$ encodes the text sequence into embeddings and incorporates visual features 
to generate the reasoning sequence 
$\mathcal{X} = (x_1, x_2, \ldots, x_N)$.
 \vspace{-0.3em}
\begin{equation}
\pi_\theta(\mathcal{X} \mid q, z)
= \prod_{n=1}^{N} \pi_\theta(x_n \mid \mathcal{X}_{<n}, q, z)
\vspace{-0cm}
\label{eq:gen}
\end{equation}
where $x_{<n}$ denotes the sequence of tokens preceding position $n$. Different from approaches that use the last hidden state of the previous reasoning step as latent think tokens~\cite{chen2025reasoningdarkinterleavedvisiontext, pham2025multimodalchaincontinuousthought}, we introduce $L$ learnable \textit{latent think tokens} into the input sequence, whose embeddings after projection are denoted as $\mathcal{T} = [\,\boldsymbol{\tau}_1,\, \boldsymbol{\tau}_2,\, \dots,\, \boldsymbol{\tau}_L\,]$. These tokens are concatenated with the original inputs and fed into the model. During test-time inference, our core idea is to keep model parameters fixed and improve reasoning solely by optimizing the embeddings of the latent think tokens. Motivated by the observations in Section~\ref{confidence}, we define a reward function $\mathcal{R}$ to quantify the confidence of the current latent reasoning state. This leads to the following test-time optimization objective:
 \vspace{-0.3em}
\begin{equation}
\mathcal{T}^* = \arg\max_{\mathcal{T}} \ \mathcal{R}(\mathcal{T},\, \mathcal{Q}, \mathcal{Z}),
\vspace{-0cm}
\end{equation}
In practice, the model iteratively update the latent think tokens for $T$ steps, allowing them to progressively evolve toward directions that maximize the reward.

\subsection{Dynamic Multimodal Latent Reasoning}

\begin{wrapfigure}{r}{0.5\linewidth}
\vspace{-4em}
    \centering
    \includegraphics[width=\linewidth]{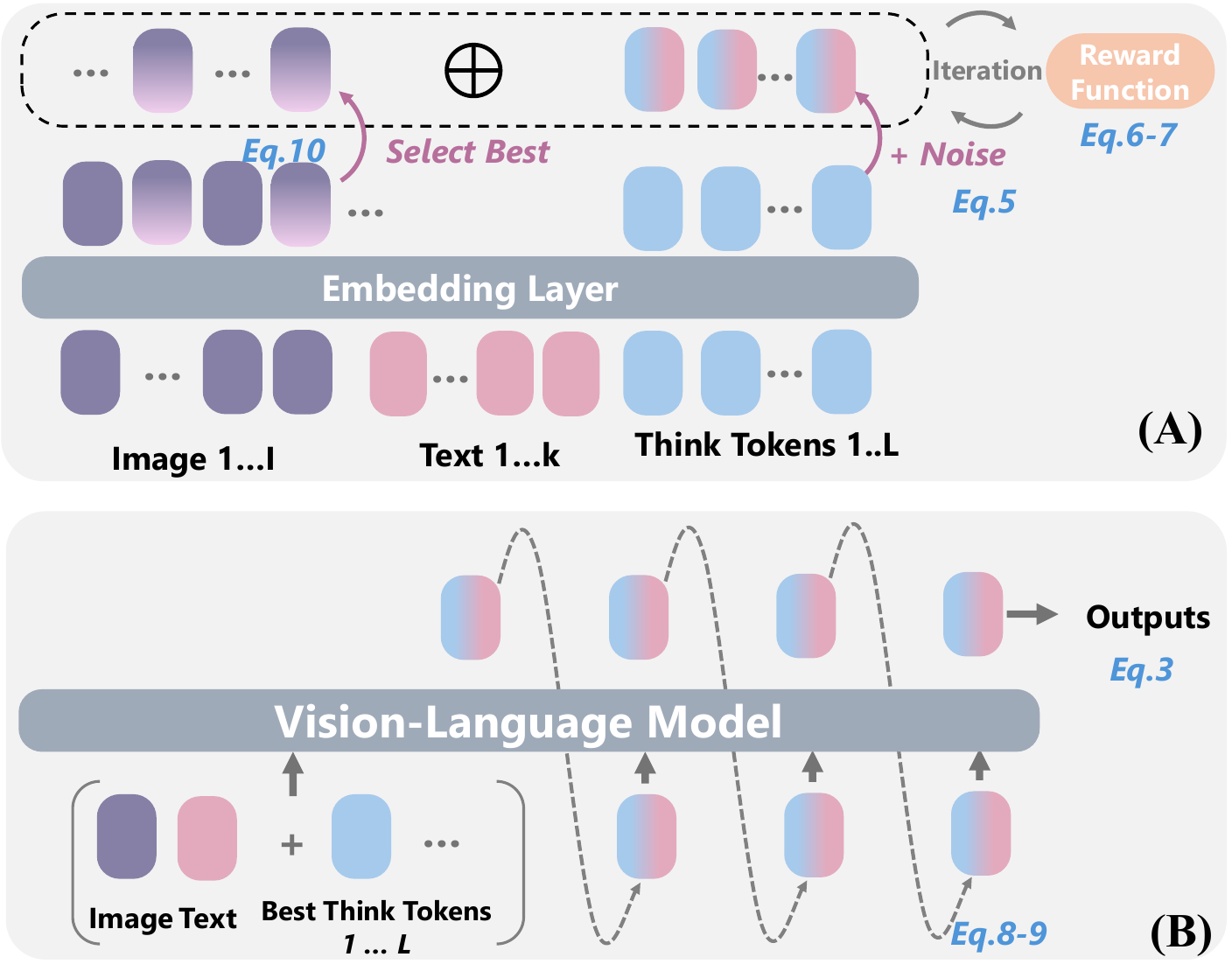}
    \caption{\small
    Overview of the proposed DMLR framework. 
    The model performs exploration through controlled noise (Eq.~\ref{noise}) and iteratively optimizes the latent think tokens via confidence-guided policy updates (Eq.~\ref{eq:policy-update1}--\ref{eq:policy-update2}).
    Dynamic Visual Injection (Eq.~\ref{concat}) selects and updates the best visual patches during optimization, and the optimized latent tokens are decoded (Eq.~\ref{eq:gen}) to produce the output.}
    \label{method}
\vspace{-2.5em}
\end{wrapfigure}

In light of the observations in Section~\ref{motivation}, DMLR comprises two key processes: dynamic visual injection strategy for \textbf{\textit{RQ1}}, and confidence-guided optimization of latent think tokens for \textbf{\textit{RQ2}}, as shown in Figure~\ref{method} and Algorithm~\ref{latent_reasoning}.

\noindent\textbf{Latent Think Tokens Initialization.} 
We initialize the latent think tokens before each iteration to facilitate exploration in the latent space. To this end, we adopt a stochastic perturbation strategy that adds controlled randomness while preserving representation stability. Specifically, multiplicative noise sampled from a Gaussian distribution is applied as a local perturbation to the current latent state:

\begin{equation}
\mathcal{T}'^{(t)} = \mathcal{T}^{(t)}  + \xi^{(t)}, \quad 
\xi^{(t)} \sim \mathcal{N}(0, \sigma^2 I)
\label{noise}
\end{equation}
where $\sigma^{2}$ is a variance hyperparameter that controls the magnitude of exploration and $\xi^{(t)}$ is the multiplicative Gaussian noise sampled at iteration $t$. More analyses and results are shown in Section~\ref{ablation}.

\noindent\textbf{Reward Formulation.} We propose a confidence-guided reward that dynamically optimizes latent think tokens during reasoning. In contrast to prior approaches~\cite{zhi2025seeingreasoningconfidencesupercharging, zou2025looktwiceanswermemoryspace} that use confidence ony for post-hoc evaluation, we treats it as an intrinsic feedback signal that continuously guides latent reasoning optimization. Given the latent think state $\mathcal{T}^{(t)}$, the query $q$, and visual features $z$, the model $\pi_{\theta}$ generates token-level probability distributions $\mathcal{P}_i^{(t)}$ over the vocabulary $w$. We further quantify the model’s confidence for each latent think token by computing the truncated entropy over its top-$k$ most probable tokens, defined as:

\begin{equation}
\mathcal{H}_k\bigl(\mathcal{P}_i^{(t)}\bigr)
= - \sum_{w \in \operatorname{Top}_k\bigl(\mathcal{P}_i^{(t)}\bigr)}
\mathcal{P}_i^{(t)}(w)\,\log\bigl(\mathcal{P}_i^{(t)}(w)\bigr)
\end{equation}
where $\mathrm{Top}_k(\cdot)$ denotes the set of the $k$ tokens with the highest probabilities. A lower value of the entropy $\mathcal{H}_k(\cdot)$ corresponds to higher confidence in the model’s prediction at that position. The reward for the entire latent reasoning sequence is defined as the complement of the mean truncated entropy computed over all  $L$ latent think tokens:

\vspace{-0.3cm}
\begin{equation}
\mathcal{R}(\mathcal{T}^{(t)}) = 1 - \frac{1}{L} \sum_{i=1}^L \mathcal{H}_k(\mathcal{P}_i^{(t)})
\label{eq:joint_reward_final}
\vspace{-0.2cm}
\end{equation}

\noindent\textbf{Test-Time Latent Optimization.}  Recent works~\cite{li2025seekdarkreasoningtesttime, zhang2025latentevolveselfevolvingtesttimescaling, ye2025thinkingflytesttimereasoning} have explored test-time gradient optimization to enable adaptation in language tasks, whereas we focus on optimization processes for multimodal latent reasoning. Specifically, during the test-time inference, guided by the objective defined in Equation~\ref{eq:joint_reward_final}, we adopt a \textit{REINFORCE-based}~\cite{williams1992simple} direct policy gradient method 
to adaptively optimize the latent think tokens $\mathcal{T}^{(t)}$. 
Assuming that each latent think token is independent, 
the update rule is formulated as:
\begin{equation}
\mathcal{T}^{(t)} \leftarrow \mathcal{T}^{(t)} + \eta \nabla_{\mathcal{T}^{(t)}} \mathcal{J}(\mathcal{T}^{(t)})
\label{eq:policy-update1}
\end{equation}
where $\eta$ denotes the learning rate. According to the Policy Gradient Theorem and Equation~\ref{noise}, the gradient can be formulated and further expressed as:
\begin{equation}
\nabla_{\!\mathcal{T}}\mathcal{J}(\mathcal{T})
= \mathbb{E}_{\mathcal{T}' \sim \pi(\cdot|\mathcal{T})}
\!\left[ \mathcal{R}(\mathcal{T}')\,\nabla_{\!\mathcal{T}}\log \pi(\mathcal{T}' \mid \mathcal{T}) \right]
= \mathbb{E}\!\left[ \mathcal{R}(\mathcal{T}')\,\frac{\xi}{\sigma^{2}} \right].
\label{eq:policy-update2}
\end{equation}

\noindent \textbf{Visual Injection Strategy.} 
Different from methods that directly inject high-attention regions~\cite{gao2025interleavedmodalchainofthought}, our strategy updates the most informative visual patches based on the reward at each iteration and injects them as latent visual tokens. As illustrated in Algorithm~\ref{latent_reasoning}, we first use the initial attention of the latent think token to collect $m$ highly relevant image patches (see Section~\ref{par}), which serve as the initial best patch $\mathcal{V}_{\textit{best}}$. At each iteration, the model resamples $m$ candidate patches 
$\mathcal{Z}_{\textit{cand}}=\{\mathcal{Z}_1,\ldots,\mathcal{Z}_m\}$ 
based on the updated attention and injects them together with the previous best patch into the latent sequence for reward, as formulated in Equation~\ref{concat}. If the reward $r > r_{\textit{best}}$, indicating that the candidate patches provide enhanced visual evidence, the best patch $\mathcal{V}_{\text{best}}$ is updated; otherwise, the previous best is retained. 
\begin{equation}
\vspace{-0.15cm}
r = \mathcal{R}\Big(\mathcal{Z}, \mathcal{Q},[\mathcal{T}^{(t)}, \mathcal{V}_{\textit{best}}, \mathcal{Z}_{\textit{cand}}]\Big)
\label{concat}
\end{equation}
As the iterations progress, the best visual patch converges to the regions most relevant to the latent think state, guiding the latent reasoning toward more effective optimization.

\begin{algorithm}[t] 
\caption{\small \textbf{\textit{Dynamic Multimodal Latent Reasoning}}}
\small
\label{latent_reasoning}
\DontPrintSemicolon
\SetAlgoVlined
\SetKwProg{Fn}{}{}{}
\KwRequire{Image embeddings $\mathcal{Z}$, text embeddings $\mathcal{Q}$, 
latent tokens $\mathcal{T}_l$, learning rate $\eta$, 
iterations $T$, best visual patch $\mathcal{V}_{\textit{best}}$, 
top-$k$ probability $\mathrm{Top}_k(\mathcal{P}_i)$, the number of candidate patches 
$m$}

\vspace{2pt}
$\mathrm{Top}_k(\mathcal{P}_i) = \pi_{\theta}([\mathcal{Q}, \mathcal{Z}, \mathcal{T}])$; 
$r \gets \mathcal{R}(\mathcal{P}_{i})$ 
\hspace{1.em}\textcolor{blue}{$\triangleright$ reward}

\vspace{3pt}
\textcolor{darkgray}{\# \textit{Latent Policy Gradient Optimization}}\;

\For{$T = 1 \dots t$}{
    $\epsilon \sim \mathcal{N}(0,\sigma^2 I)$ 
    \hfill \textcolor{blue}{$\triangleright$ latent perturbation}

    $\mathcal{T}^{(t)\prime} \gets \mathcal{T}^{(t)} + \epsilon$

    $\mathcal{T}^{(t)} \gets \mathcal{T}^{(t)} + 
    \eta \nabla_{\mathcal{T}^{(t)}} \mathcal{J}(\mathcal{T}^{(t)})$
    \hfill \textcolor{blue}{$\triangleright$ latent update}

    \vspace{2pt}
    \textcolor{darkgray}{\# \textit{Dynamic Visual Injection}}\;
    
    $\mathcal{V}_{\textit{best}} \gets \text{Initialize}(\mathcal{T}^{(0)}, m)$ \hfill \textcolor{blue}{$\triangleright$ initialize best patch}

    \For{$L = 1 \dots l$}{
    
       $\mathcal{Z}_{\textit{cand}} \gets 
        \text{AttentionSelect}(\mathcal{T}_l^{(t)}, m)$
        \hfill \textcolor{blue}{$\triangleright$ select $m$ candidate visual patches}

        $\tilde{\mathcal{T}_l}^{(t)} \gets [\mathcal{T}_l^{(t)}, \mathcal{Z}_{\textit{cand}}, \mathcal{V}_{\textit{best}}]$

        $r \gets \mathcal{R}(\mathcal{Q}, \mathcal{Z}, \tilde{\mathcal{T}^{(t)}_l})$

        \If{$r > r_{\text{best}}$}{
            $\mathcal{V}_{\textit{best}} \gets \mathcal{V}_{\textit{best}} \cup \mathcal{Z}_{\textit{cand}}$;

            $\mathcal{T}_l^{(t)} \gets \tilde{\mathcal{T}_l}^{(t)}$
            \hfill \textcolor{blue}{$\triangleright$ update best}
        }
        \Else{
            $\mathcal{T}_l^{(t)} \gets 
            [\mathcal{T}_l^{(t)}, \mathcal{V}_{\textit{best}}]$
            \hfill \textcolor{blue}{$\triangleright$ revert to previous best}
        }
        \vspace{1pt}
    }
    \vspace{2pt}
}

\vspace{3pt}
$\mathcal{X} \gets \text{Decode}(\mathcal{T}^{(t)}, \mathcal{Z}, \mathcal{Q})$

\Return $\mathcal{X}$
\end{algorithm}

\subsection{Theoretical Analysis}
To further understand why DMLR achieves high efficiency and robust performance, we provide theoretical explanations through the following two theorems. \\
\noindent \textbf{Theorem 4.1} (Confidence Reflects Reasoning Quality)\textbf{.} \textit{Let \(h\) denote the latent reasoning state in DMLR, where \(C(h)\) represents the model’s confidence level 
and \(Q(h)\) denotes the corresponding reasoning quality. If and only if the gradients of \(C(h)\) and \(Q(h)\) are positively aligned, 
the DMLR update along the confidence ascent direction will consequently improve the reasoning quality:
}  
\normalsize
\begin{equation}
\nabla C(h) \cdot \nabla Q(h) > 0
\end{equation}

\noindent \textbf{Theorem 4.2.} (Visual Injection Enhances Confidence)\textbf{.}
\textit{Let $\mathcal{T}$ be the latent reasoning states, $\hat{\mathcal{T}}$ denote the updated states after visual injection, and $z_v$ be the visual features. Visual injection in DMLR increases the mutual information between latent states and visual features, thereby enhancing the expected confidence $J_{\text{conf}}(\mathcal{T})$, satisfying:}
\begin{equation}
I(\hat{\mathcal{T}}; z_v) \ge I(\mathcal{T}; z_v) \ \Rightarrow\ 
J_{\text{conf}}(\hat{\mathcal{T}}) \ge J_{\text{conf}}(\mathcal{T})
\end{equation}

\section{Experiments}

\subsection{Experiment Setup}
\label{setup}
\noindent\textbf{Baselines.} We evaluate the proposed DMLR  using two types of baselines: model-based and method-based. For the model baselines, we consider six representative MLLMs, including two reasoning models, R1-OneVision~\cite{yang2025r1onevisionadvancinggeneralizedmultimodal} and VLAA-Thinking~\cite{chen2025sftrlearlyinvestigation}, as well as four non-reasoning models, Qwen2.5-VL-3B/7B~\cite{bai2025qwen25vltechnicalreport} and Qwen3-VL-4B/8B~\cite{qwen3vl2025}. For method baselines, we consider two reasoning paradigms: \textit{Text-only Reasoning} (CCoT~\cite{Mitra_2024_CVPR}) and \textit{Vision-Text Involved Reasoning} (ICoT~\cite{gao2025interleavedmodalchainofthought}, Multimodal-CoT~\cite{zhang2024multimodalchainofthoughtreasoninglanguage}). We further include a Vanilla baseline where non-reasoning models answer directly and reasoning models use their default prompts.

\noindent\textbf{Evaluation Benchmarks.} We evaluate our method on three tasks across six benchmarks: (1) \textit{Mathematics Reasoning} (MathVista$_{\text{mini}}$~\cite{lu2024mathvistaevaluatingmathematicalreasoning}, MathVision$_{\text{mini}}$~\cite{wang2024measuringmultimodalmathematicalreasoning}, MM Math~\cite{sun2024mmmathadvancingmultimodalmath} ); (2) \textit{Visual Reasoning} (HallusionBench~\cite{guan2024hallusionbenchadvanceddiagnosticsuite}, MMVP~\cite{tong2024eyeswideshutexploring}); (3) \textit{Multimodal Composition} (MMStar~\cite{chen2024rightwayevaluatinglarge}, ScienceQA~\cite{lu2022learnexplainmultimodalreasoning}). Details are provided in Appendix~\ref{app:eval}.

\noindent\textbf{Implementation Details.} \label{par} All frameworks adopt the eager attention mode to enable access to internal attention maps. A total of 4 latent think tokens $\mathcal{T}$ are used, with $m = 2$ visual candidate patches injected at each iteration.
The defaultnumber of optimization iterations is set to 15, with a learning rate of $10^{-3}$. To ensure stable exploration in the latent space, the perturbation magnitude  $\sigma$ is set to $10\%$. All experiments are conducted on four NVIDIA H100 GPUs, with further detailed parameter analyses in Appendix~\ref{app:parameter}.

\subsection{Main Results}

\begin{table*}[t]
\centering
\caption{Comparison of different reasoning methods and DMLR across various benchmarks. All metrics are reported in Accuracy (\%).
Results are evaluated over a diverse suite of mathematics reasoning, visual reasoning, and multimodal composition tasks under multiple backbone models.}

\large
\renewcommand{\arraystretch}{1.05}
\resizebox{0.9\linewidth}{!}{%
\setlength{\tabcolsep}{3pt}
\begin{tabular}{ll@{\hspace{6pt}}ccc@{\hspace{6pt}}cc@{\hspace{6pt}}cc}
\toprule[1.5pt]
\multicolumn{2}{c}{} &
\multicolumn{3}{c}{\textbf{Mathematics Reasoning $\uparrow$}} &
\multicolumn{2}{c}{\textbf{Visual Reasoning $\uparrow$}} &
\multicolumn{2}{c}{\textbf{Multimodal Composition $\uparrow$}} \\
\cmidrule(lr){3-5} \cmidrule(lr){6-7} \cmidrule(lr){8-9}
\textbf{Method} & \textbf{Model} &
$\mathrm{MathVista}_{\text{mini}}$ &
$\mathrm{MathVision}_{\text{mini}}$ &
MM-Math &
HallusionBench &
MMVP & MMStar &
ScienceQA \\
\midrule

Vanilla &  & 58.7 & 21.6 & 37.5 & 65.4 & 68.7 & 59.3 & 49.7 \\
Multimodal COT &  & 56.4 & 21.8 & 35.6 & 63.6 & 68.1 & 57.9 & 49.5  \\
CCOT &  & 57.8 & 22.5 & 36.3 & 64.9 & 69.0 & 58.7 & 50.2 \\
ICoT &  & 58.9 & 23.3 & 37.0 & 65.5 & 69.3 & 60.4 & 50.4\\
\rowcolor{lightblue}
\textbf{\textit{+DMLR (Ours)}} &
\multirow[c]{-5}{*}[1pt]{\rotatebox{90}{\parbox{2.5cm}{\centering \fontsize{11pt}{13pt}\selectfont Qwen2.5 \\ VL 7B}}}
 & \textbf{59.1} \up{0.40\%} & \textbf{24.4} \up{2.8\%} & 38.8 \up{1.3\%} & 65.8 \up{0.4\%} & \textbf{70.1} \up{1.4\%} & \textbf{60.1} \up{0.8\%} & \textbf{51.3} \up{1.6\%} \\
\midrule\midrule

Vanilla &  & 48.2 & 15.7 & 29.0 & 64.2 & 55.6 & 50.2 & 44.1 \\
Multimodal COT &  & 47.3 & 14.3 & 28.5 & 63.8 & 54.4 & 48.5 & 42.9 \\
CCOT & & 48.0 & 15.6 & 30.2 & 64.0 & 55.5 & 49.3 & 44.5 \\
ICoT &  & 49.8 & 16.0 & 30.6 & 64.7 & 55.9 & 49.6 & 45.0 \\
\rowcolor{lightblue}
\textbf{\textit{+DMLR (Ours)}} &
\multirow[c]{-5}{*}[1pt]{\rotatebox{90}{\parbox{2.5cm}{\centering \fontsize{11pt}{13pt}\selectfont Qwen2.5 \\ VL 3B}}}
 & \textbf{51.0} \up{2.80\%} & \textbf{17.7} \up{2.74\%} & \textbf{33.3} \up{4.3\%} & \textbf{64.7} \up{0.5\%} & \textbf{56.8} \up{1.26\%} & \textbf{51.2} \up{1.00\%} & \textbf{46.9} \up{2.8\%}\\
\midrule\midrule

Vanilla &  & 61.1 & 23.5 & 41.5 & 62.0 & 68.3 & 58.9 & 50.6 \\\
Multimodal COT &  & 59.6 & 23.1 & 40.6 & 62.8  & 67.2 & 57.1 & 48.2  \\
CCOT &  & 60.5  & 24.8  & 41.8  & 64.6 & 68.0 & 59.0 & 49.4\\
ICoT &  & 61.4 & 25.0 & 42.3 & 65.9 & 68.3 & 58.2 & 50.6 \\
\rowcolor{lightblue}
\textbf{\textit{+DMLR (Ours)}} &
\multirow[c]{-4}{*}[1pt]{\rotatebox{90}{\parbox{1.0cm}{\centering \fontsize{11pt}{13pt}\selectfont VLAA \\ Thinking 7B}}}
 & \textbf{62.9} \up{1.80\%} & \textbf{27.6} \up{4.10\%} & \textbf{43.9} \up{2.41\%} & \textbf{67.9} \up{5.94\%} & \textbf{69.4} \up{1.1\%} & \textbf{59.2} \up{0.3\%} & \textbf{51.98} \up{1.38\%} \\
\midrule\midrule

Vanilla &  & 51.2 & 18.7 & 40.7 & 62.1 & 67.0 & 52.1 & 50.9 \\
Multimodal COT &  & 52.5 & 18.9 & 39.6 & 62.5 & 68.0 &  51.6 & 51.7 \\
CCOT &  & 53.4 & 20.3 & 40.8 & 63.0 & 68.9 & 53.5 & 52.8 \\
ICoT &  & 55.6  & 21.5 & 41.7 & 63.8 & 69.6 & 54.0 & 54.4 \\
\rowcolor{lightblue}
\textbf{\textit{+DMLR (Ours)}} &
\multirow[c]{-4}{*}[1pt]{\rotatebox{90}{\parbox{1.2cm}{\centering \fontsize{11pt}{11pt}\selectfont \quad R1 \\ OneVision \quad \quad 7B}}}
 & \textbf{58.0} \up{6.81\%} & \textbf{23.3} \up{4.56\%} & \textbf{42.9} \up{2.21\%} & \textbf{64.1} \up{2.09\%} & \textbf{71.9} \up{4.93\%} & \textbf{56.2} \up{4.14\%} & \textbf{55.4} \up{4.52\%}\\
\midrule\midrule

Vanilla &  & 66.0 & 32.9 & 66.2 & 73.2 & 71.9 & 68.1 & 54.1 \\
Multimodal COT &  & 64.8 & 32.8 & 65.1 & 73.0 & 69.6 & 66.9  &  53.2 \\
CCOT &   & 66.5 & 33.3 & 65.5 & 73.5 & 70.3 & 68.8 & 54.4 \\
ICoT &  & 66.2 & 34.9 & 66.8 & 74.5 & 71.8 & 69.3 & 55.8 \\
\rowcolor{lightblue}
\textbf{\textit{+DMLR (Ours)}} &
\multirow[c]{-5}{*}[1pt]{\rotatebox{90}{\parbox{2.5cm}{\centering \fontsize{11pt}{13pt}\selectfont Qwen3 \\ VL 8B}}}
 & \textbf{66.9} \up{0.9\%} & \textbf{36.2} \up{3.34\%}& \textbf{67.7} \up{1.51\%} & \textbf{74.6} \up{1.48\%} & \textbf{72.8} \up{0.97\%}& \textbf{70.0} \up{1.91\%} & \textbf{55.6} \up{1.48\%} \\
\midrule\midrule

Vanilla &  & 64.7 & 24.2 & 65.4 & 71.6 & 71.3 & 57.4 & 52.4 \\
Multimodal COT &  & 62.3 & 24.8 & 63.9 & 70.0 & 69.6 & 57.7 & 53.0  \\
CCOT &  & 64.5 &  26.6 & 64.8 & 71.5 & 71.2 & 58.8 & 52.9 \\
ICoT &  & 64.5 & 27.5 & 65.0 & 72.2 & 72.5 & 59.3 & 53.7 \\
\rowcolor{lightblue}
\textbf{\textit{+DMLR (Ours)}} &
\multirow[c]{-5}{*}[1pt]{\rotatebox{90}{\parbox{2.5cm}{\centering \fontsize{11pt}{13pt}\selectfont Qwen3 \\ VL 4B}}}
 & \textbf{65.6} \up{0.93\%} & \textbf{29.4} \up{5.20\%}  & \textbf{65.9} \up{0.5\%} & \textbf{72.7} \up{1.12\%} & \textbf{72.3} \up{0.98\%}& \textbf{60.3} \up{2.88\%} & \textbf{54.9} \up{2.48\%} \\
\bottomrule[1.5pt]
\end{tabular}
}
\vspace{-0.5em}
\label{tab:baseline}
\end{table*}

\textbf{Overall Results.}  As shown in Table~\ref{tab:baseline}, models integrated with DMLR achieve the best performance on over 95\% of tasks. On mathematical and visual reasoning benchmarks, Qwen2.5-VL-7B achieves average improvements of +1.5\% in mathematics and +0.9\% in visual reasoning, while the reasoning counterpart R1-OneVision attains average gains of +4.5\% and +3.45\% on the two domains, respectively. These results indicate that DMLR generalizes robustly across diverse model paradigms and scales. Unlike other baseline methods that often involve trade-offs between reasoning and perception, DMLR consistently improves performance in both domains. For instance, while ICoT yields noticeable gains on mathematical tasks but provides only limited improvements on visual reasoning (e.g., MMVP), DMLR delivers more stable cross domain enhancements, with DMLR-integrated VLAA-Thinking averaging +2.43\% higher across all benchmarks.

\subsection{Ablation Study}
\label{ablation}

\begin{table}[H]
\vspace{-0.2cm}
\centering
\setlength{\tabcolsep}{2pt}
\caption{\small Ablation on Latent Visual Injection. We compare different injection strategies across multiple benchmarks. 
\textit{All} injects all visual patches at every iteration, while \textit{Ours} injects the best visual patches. Refer to Section~\ref{setup} for detailed settings.}

\resizebox{0.5\linewidth}{!}{
\begin{tabular}{lcccc}
\bottomrule
\textbf{Method} & MathVista & MathVision & MMStar & ScienceQA \\
\midrule\midrule
w/o Injection & 0.627  & 0.321  & 0.687  &  0.536 \\
+ Injection (All) & 0.621 & 0.327  & 0.676  & 0.527  \\
\rowcolor{lightblue} + \textbf{DVI (Ours)} &  \textbf{0.634}  & \textbf{0.340} & \textbf{0.694}  & \textbf{0.549}  \\
\bottomrule
\end{tabular}
}
\vspace{-0.3cm}
\label{tab:visual_ablation}
\end{table}

\noindent\textbf{Impact of Visual Injection Strategies.} We evaluate various visual injection strategies to assess their effects on reasoning performance. As shown in Table~\ref{tab:visual_ablation}, removing visual injection maintains stable reasoning results but leads to a clear drop in perceptual accuracy, underscoring the necessity of visual cues during latent optimization. Injecting all visual patches enhances perception but introduces instability due to redundant visual information. In contrast, DMLR exhibits consistently more stable performance, indicating that its continuously selects more relevant and stable visual information throughout the iterative optimization.

\begin{figure}[h]
\centering
\begin{minipage}{0.48\linewidth}
    \centering
    \includegraphics[width=\linewidth]{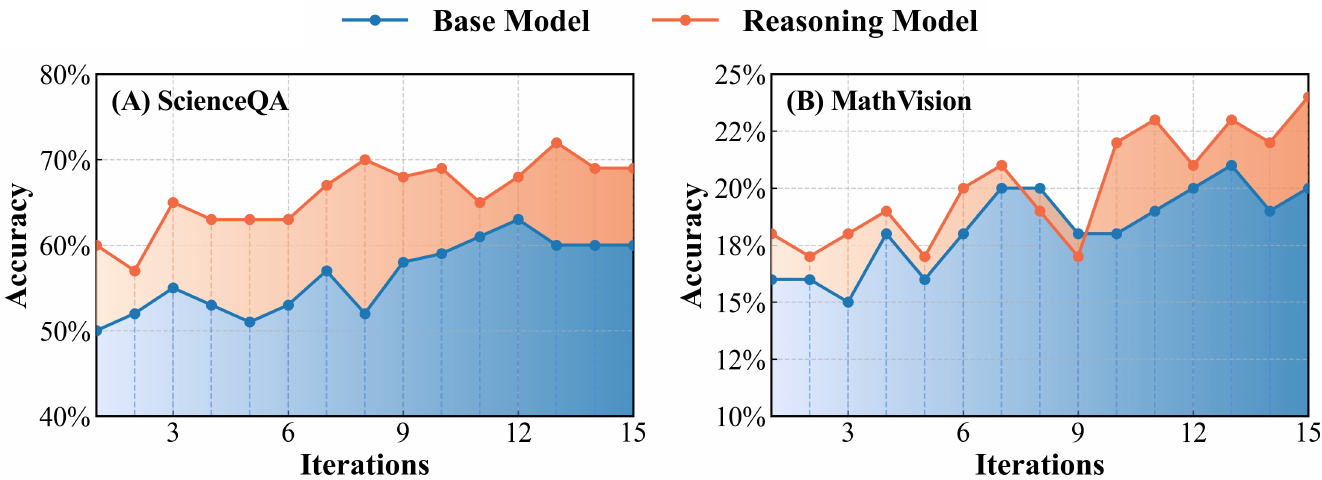}
    \captionof{figure}{\small
    Effect of iterations on performance. For both the base model and the reasoning model, accuracy on both datasets increases as the number of iterations grows.
    }
    \label{fig:acc_ite}
\end{minipage}
\hfill
\begin{minipage}{0.48\linewidth}
    \centering
    \includegraphics[width=\linewidth]{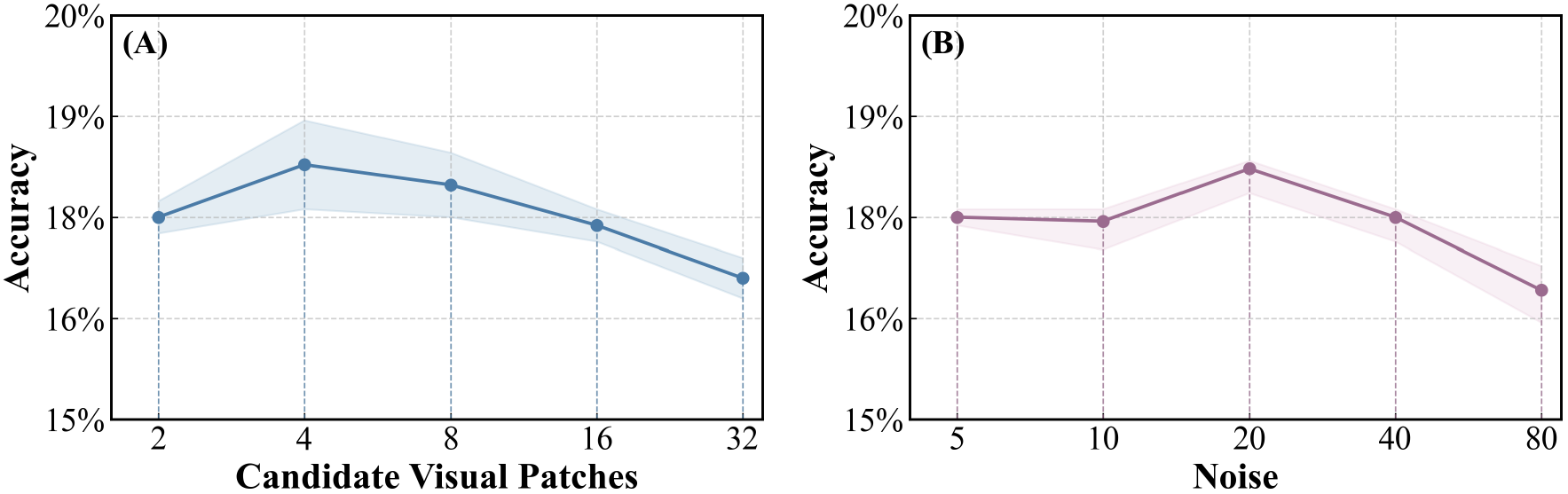}
    \captionof{figure}{\small
    (A) Effect of the number of injected candidate visual patches on performance. (B) Impact of noise magnitude (\%) on performance. All results are evaluated on the MathVision dataset.
    }
    \label{fig:patch_noise}
\end{minipage}
\end{figure}

\noindent\textbf{Impact of Iteration Number.} As shown in Figure~\ref{fig:acc_ite}, increasing the number of iterations leads to a steady improvement on both reasoning and perception tasks, indicating that iterative optimization effectively enhances latent reasoning. Morever, the reasoning model maintains consistently higher accuracy throughout the process and continues to yield gains even after multiple iterations, demonstrating a stronger ability to benefit from iterative refinement.

\noindent\textbf{Impact of Noise Scale.}
We further analyze the influence of the perturbation magnitude $\sigma$ on latent optimization. As shown in Figure~\ref{fig:patch_noise}(b), increasing the initial noise scale promotes effective exploration, allowing the model to cover a wider range of latent trajectories and identify higher-confidence reasoning paths. However, when  $\sigma$ becomes excessively large, the injected perturbation makes the updates unstable, leading to a subsequent drop in performance. This indicates that latent reasoning benefits from only a modest level of perturbation.

\begin{figure}[h]
\centering
\begin{minipage}{0.48\linewidth}
    \centering
    \includegraphics[width=\linewidth]{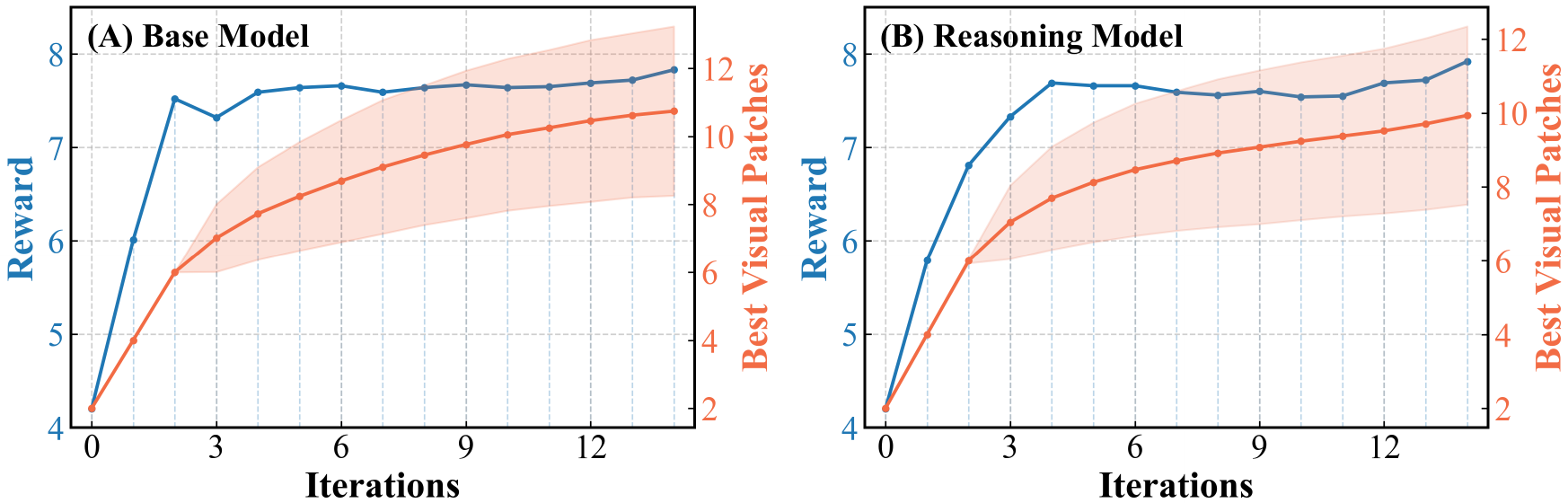}
    \captionof{figure}{\small
    Confidence reward and best visual patch injection across iterations. Both the base model and the reasoning model exhibit a clear positive correlation.
    }
    \label{fig:reward_patches}
\end{minipage}
\hfill
\begin{minipage}{0.48\linewidth}
    \centering
    \includegraphics[width=\linewidth]{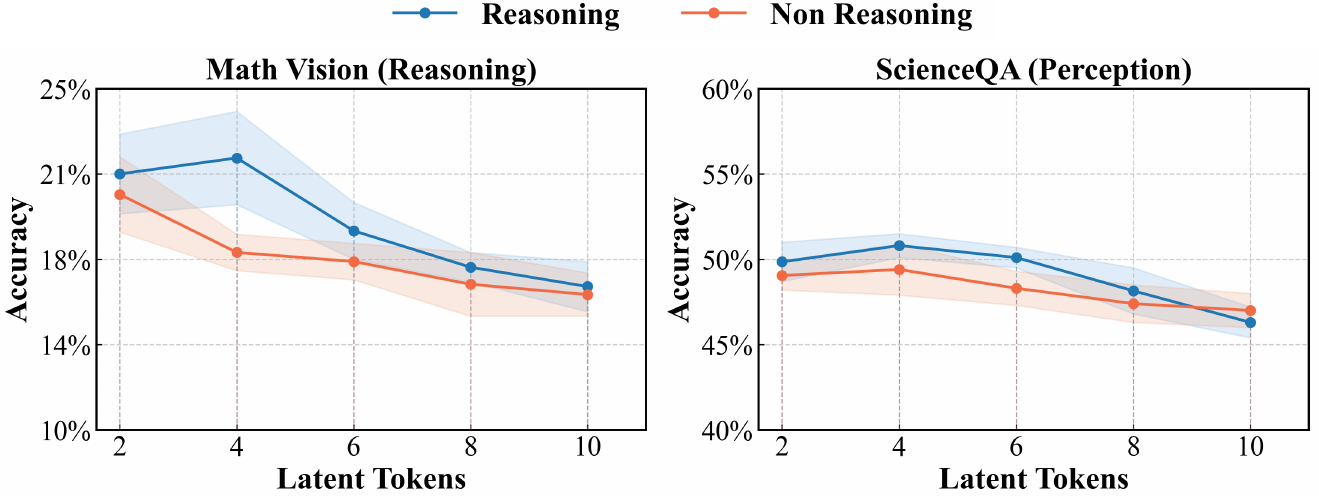}
    \captionof{figure}{\small
    Effect of the number of latent tokens. Increasing the number of latent tokens initially improves performance, but excessive tokens lead to noticeable degradation.
    }
    \label{fig:acc_token}
\end{minipage}
\end{figure}

\noindent\textbf{Impact of Visual Patch Number.}  As shown in Figure~\ref{fig:patch_noise}(a), performance improves when a moderate number of candidate visual patches are injected, whereas injecting an excessive number of patches leads to a clear decline. This trend indicates that a limited number of candidates is sufficient for effective updates, while excessive patches introduce redundant visual information that negatively affect optimization. Furthermore, Figure~\ref{fig:reward_patches} shows that as the iterations progress, the reward steadily increases and the selected best patch becomes increasingly stable, exhibiting a clear convergence trend. This trend indicates that the dynamic injection strategy does not continually introduce additional visual patches into the latent space, but instead converges toward a small set of highly relevant patches during optimization.

\noindent\textbf{Number of Latent Think Tokens.}
We further evaluate the impact of the number of latent think tokens on overall performance. As shown in Figure~\ref{fig:acc_token}, setting the number of latent tokens to a small range (2–4) yields stable improvements on both reasoning and perception tasks. However, as the number of tokens continues to increase, performance on both tasks begins to decline, with the reasoning model exhibiting more pronounced fluctuations. This overall trend indicates that increasing the number of latent tokens beyond a moderate level does not provide additional benefits and instead makes the optimization process less stable.

\begin{figure*}[t]
\centering
\includegraphics[width=1\textwidth]{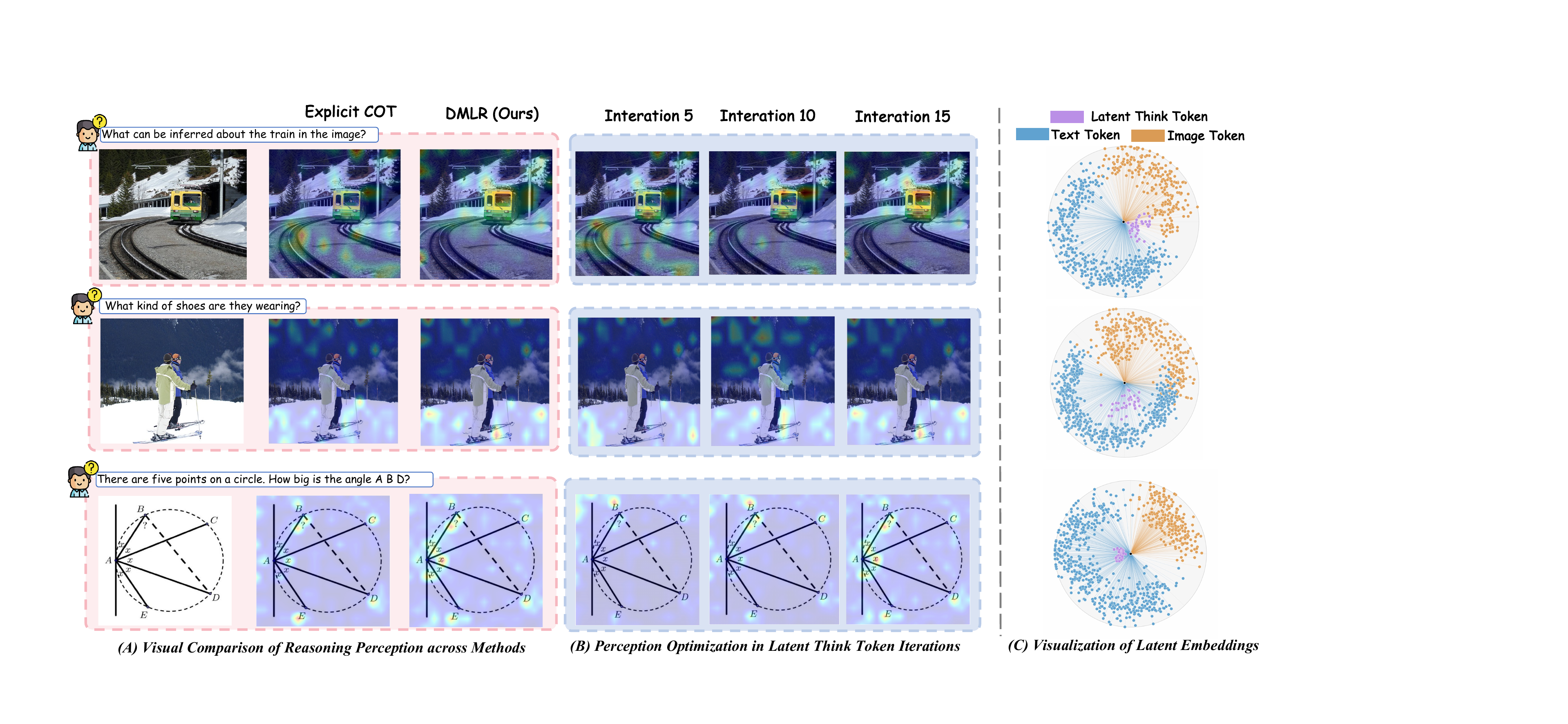}

\caption{
Qualitative analysis of our DMLR framework. 
(A) Visual comparison of visual grounding behaviors between Explicit CoT and DMLR across diverse queries. DMLR produces more focused and stable visual grounding than explicit CoT. 
(B) Perception optimization across latent think token iterations, where visual attention becomes progressively sharper and better aligned with relevant regions. 
(C) Visualization of latent embeddings showing the geometric separation of latent think tokens, text tokens, and image tokens, illustrating the structured organization of the latent reasoning space.
}
\label{fig:perception}

\end{figure*}

\subsection{Quantitative Analysis}
\noindent\textbf{Visual Grounding Analysis.} 
We visualize the attention heatmaps of VLAA-Thinking during the reasoning process. As shown in Figure~\ref{fig:perception}(a), the explicit CoT baseline often shifts its attention toward task-irrelevant regions, whereas DMLR maintains a stable focus on task-relevant areas. This demonstrates that latent multimodal reasoning produces more consistent and reliable visual grounding throughout the reasoning process. Figure~\ref{fig:perception}(b) further shows the evolution of attention across iterations. The attention distribution gradually converges toward task-relevant regions in models integrated with DMLR, reflecting a more stable and consistent focus throughout the optimization.

\noindent\textbf{Latent Behavior Analysis.}
We visualize the final distributions of latent think tokens, text tokens, and image tokens using t-SNE \cite{maaten2008visualizing} to analyze the effect of the  iterative optimization on the latent reasoning. 
As shown in Figure~\ref{fig:perception}(c), the latent think tokens form a tight cluster that is well separated from both text and visual embeddings, and are located in a stable intermediate region between the two modalities.  This distribution suggests that the optimized latent tokens become modality-independent, forming a unified cross-modal semantic representation. The compactness of the cluster further indicates that the optimization process yields more stable and consistent latent reasoning states.

\begin{figure}[h]
\centering
\includegraphics[width=0.5\linewidth]{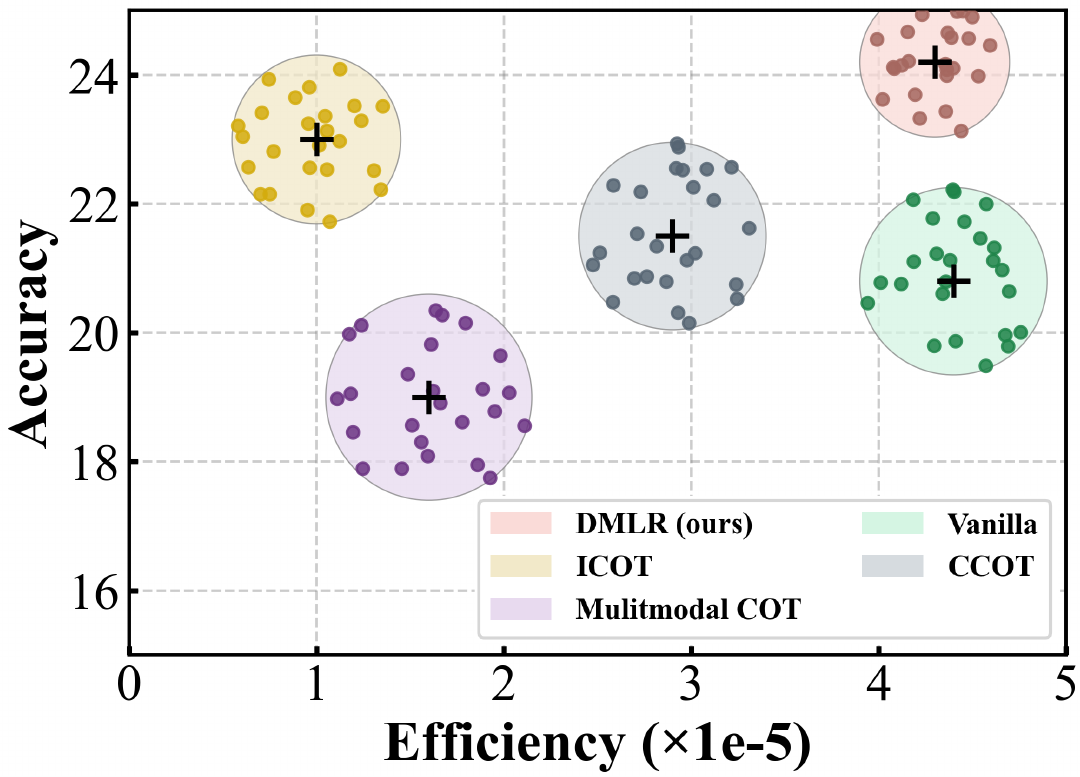}
\caption{\small
Comparison of efficiency and accuracy across various reasoning methods on the MathVision Benchmark.
DMLR achieves the best overall trade-off, delivering higher accuracy while maintaining strong inference efficiency. 
The x-axis reports the efficiency metric $(\text{Acc} / \text{AvgBatchTime})^2$.
}
\label{fig:efficiency}
\end{figure}

\noindent\textbf{Inference Efficiency Analysis.} 
As shown in Figure~\ref{fig:efficiency}, different reasoning paradigms exhibit distinct trade-offs between accuracy and efficiency. The explicit methods such as Multimodal CoT rely on long-chain text generation, incurring substantial computational overhead. Although ICoT enhances reasoning to some extent, it injects a large volume of visual information during decoding, which significantly slows inference. In contrast, DMLR performs optimization entirely within the latent space, introducing no additional sequence generation cost. Moreover, its dynamic visual injection strategy selects only the relevant visual patches to the current latent state at each iteration, eliminating redundant visual computation. By preserving accuracy gains while reducing inference overhead, DMLR achieves a more favorable balance between efficiency and performance.

\section{Conclusion}
\noindent In this work, we analyze how MLLMs utilize visual information and confidence during reasoning. Based on these observations, we introduce DMLR, a test-time multimodal latent reasoning framework that integrates confidence-guided latent optimization with dynamic visual injection.This method enables models to refine their reasoning, retrieve visual evidence only when need without training. Extensive experiments across various tasks show that DMLR consistently boosts both reasoning and perception tasks, offering a stable and training-free alternative to other methods.

\bibliographystyle{unsrtnat}  
\bibliography{ref}

\appendix
\newpage




\section*{\LARGE Appendix}

\section{More Detailed about Evaluation}

\subsection{Datasets}
\label{app:eval}

\begin{itemize}[leftmargin=*]
\item \textbf{MathVista$_{\text{mini}}$} is a benchmark for mathematical reasoning in visual contexts, aggregating diverse multimodal math tasks that require fine-grained visual understanding and compositional numerical reasoning.

\item \textbf{MathVision$_{\text{mini}}$} is a curated benchmark of competition-level visual math problems spanning multiple disciplines and difficulty levels to assess multimodal models’ mathematical reasoning under challenging and diverse settings.

\item \textbf{MM Math} is a benchmark of open-ended math problems with visual contexts that supports both outcome and process evaluation, enabling detailed analysis of multimodal reasoning behaviors and typical error patterns.

\item \textbf{HallusionBench} is a benchmark for image-context reasoning that uses carefully structured question pairs to diagnose hallucination, visual illusion, and logical inconsistency in large vision-language models.

\item \textbf{MMVP} is a benchmark built from multimodal visual patterns designed to expose “CLIP-blind’’ image–text pairs, revealing systematic visual perception failures and hallucinated explanations in multimodal LLMs. 

\item \textbf{MMStar} is a vision-indispensable multimodal benchmark composed of carefully human-filtered samples that ensure true visual dependency while evaluating core multimodal capabilities along multiple fine-grained axes.

\item \textbf{ScienceQA} is a multimodal multiple-choice science benchmark with rich textual and visual contexts, lectures, and explanations that spans diverse subjects and skills, supporting evaluation of both answer accuracy and explanation quality.
\end{itemize}

For all datasets, we limit the maximum sample size to 1000 instances.

\subsection{Evaluation Setting}
\label{app:eval_setting}
We adopt a unified prompting setup for all models. Unless otherwise stated, we use greedy decoding (\texttt{do\_sample=False}) for all generation tasks.

\paragraph{System Prompt.}
\begin{center}
\fbox{\parbox{0.96\linewidth}{
\small
A conversation between User and Assistant. The user asks a question, and the Assistant solves it. The assistant first thinks about the reasoning process in the mind and then provides the user with the answer. The reasoning process and answer are enclosed within \texttt{<think>} \texttt{</think>} and \texttt{<answer>} \texttt{</answer>} tags, respectively, i.e., \texttt{<think>} reasoning process here \texttt{</think>} \texttt{<answer>} answer here \texttt{</answer>}.
}}
\end{center}

\paragraph{Task Prompt.}
\begin{center}
\fbox{\parbox{0.96\linewidth}{
\small
Please analyze the image carefully and solve this problem step by step. Show your reasoning process clearly, then put your final answer within \texttt{\textbackslash boxed\{\}}.\\[4pt]
\textbf{Question:} \emph{[Problem Text]}
}}
\end{center}

For all benchmarks considered in our experiments, the ground-truth answers are verifiable; we use regular expressions to extract the content within \texttt{\textbackslash boxed\{\}} from the model outputs and then match it against the correct answers.

\subsection{Parameters Setup}
\label{app:parameter}

\begin{itemize}[leftmargin=*]

\item \textbf{Latent Think Tokens $\mathcal{T}$:} We set the number of latent think tokens to 4. During generation, after each latent token the model dynamically injects a visual patch into the latent stream to refresh its internal perception state.

\item \textbf{Image Patches $m$:} We dynamically insert visual patches into the latent stream. At initialization, we inject 2 patches; at each subsequent iteration, we select $m = 2$ patches with the highest attention scores and append them after each latent think token, with at most 16 patches inserted per iteration. Additionally, we set the image processor's max pixel size to 256 for all inputs.

\item \textbf{Optimization Parameters:} We perform 15 latent optimization steps with a learning rate of $1\times 10^{-3}$. To ensure stable exploration in the latent space, the perturbation magnitude is set to $\sigma = 0.1$ (10\%) with a decay factor of 0.95, and we fix the random seed to 42 for all experiments.

\item \textbf{Environment Setup:} All experiments are conducted on NVIDIA H100 GPUs. For a fair comparison, we run all baselines and our method in \texttt{float32} precision and use the \texttt{eager} attention backend for inference.
\end{itemize}

\section{Case Study}
\label{f}
In this section, we provide qualitative examples to illustrate the behavior of the DMLR-enhanced model across different task settings. As shown in the following example, the model demonstrates more reliable and fine-grained visual perception, supported by stable and semantically coherent internal representations. In reasoning scenarios, the model produces accurate and well-structured logical steps. Importantly, the dynamic latent reasoning mechanism suppresses unnecessary expansions, resulting in concise and task-focused reasoning chains rather than overly long or speculative explanations. Taken together, these examples highlight that DMLR simultaneously strengthens perceptual quality and reasoning reliability, while ensuring that the generated reasoning remains efficient and free from redundant elaboration.

\begin{figure*}[t]
    \centering
    \begin{mdframed}[
        frametitle={Perception Example},
        frametitlebackgroundcolor=lightyellow,
        backgroundcolor=gray!10,
        linecolor=gray,
        frametitlerule=true,
        frametitlefont=\color{black}\bfseries
    ]
        \centering
        \small
        \includegraphics[width=0.4\linewidth]{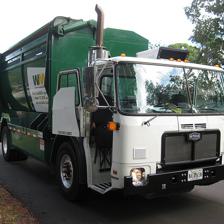} 
        \captionof{figure}{\centering Question: Is the door of the truck cab open?\\Options: (a) Yes (b) No\\Ground Truth: (a)       \label{fig1}} 
  
        \flushleft

    \noindent\textbf{Qwen2.5-VL-7B-Instruct + DMLR:}

    The image shows a garbage truck with its door on the side of the cab open. This is evident from the visible interior and the fact that the door is not aligned with the rest of the vehicle's structure, indicating it has been opened. The answer is \boxed{a}. 

    \vspace{1em}

    \noindent\textbf{Qwen2.5-VL-7B-Instruct:}

    To determine if the door of the truck cab is open, let's analyze the image step by step:
    \begin{enumerate}
        \item Observe the side of the truck where the door would be located.
        \item Look for any visible gap or opening that would indicate the door is not closed.
        \item Check the overall appearance of the door; if it is open, there should be a noticeable gap or difference in the alignment compared to when it is closed.
    \end{enumerate}

    Upon examining the image:
    \begin{itemize}
        \item The door appears to be aligned with the rest of the truck's body.
        \item There is no visible gap or difference in the alignment that would suggest the door is open.
    \end{itemize}

    Based on this analysis, the door of the truck cab does not appear to be open.

    \noindent\texttt{<think>} The door of the truck cab is not open.\texttt{</think>}
    \noindent\texttt{<answer>} No\texttt{</answer>}

    \noindent\boxed{\text{No}}
       \label{fig1}
    \end{mdframed}
\end{figure*}

\begin{figure*}[t]
    \centering
    \begin{mdframed}[
        frametitle={Reasoning Example},
        frametitlebackgroundcolor=lightyellow,
        backgroundcolor=gray!10,
        linecolor=gray,
        frametitlerule=true,
        frametitlefont=\color{black}\bfseries
    ]
        \centering
        \small
        \includegraphics[width=0.4\linewidth]{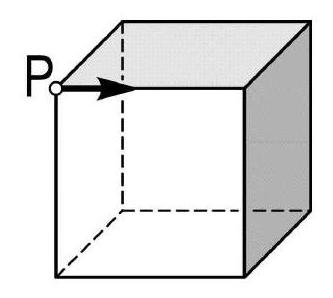} 
        \captionof{figure}{\centering Question: A beetle walks along the edges of a cube. Starting from point $P$ it first moves in the direction shown. At the end of each edge it changes the direction in which it turns, turning first right then left, then right etc. Along how many edges will it walk before it returns to point $P$?\\Ground Truth: 6} 
        \flushleft

    \noindent\textbf{Qwen2.5-VL-7B-Instruct + DMLR:}

    The beetle starts at point P and moves along the edges of a cube, changing direction according to the pattern: right, left, right. To determine how many edges it will walk before returning to point P, we need to trace its path.

    \begin{enumerate}
        \item The beetle starts at P.
        \item It moves right to an adjacent vertex.
        \item It turns left to another adjacent vertex.
        \item It turns right again to reach a third adjacent vertex.
        \item It continues this pattern until it returns to P.
    \end{enumerate}

    By examining the cube's structure, we can see that after 6 edges, the beetle will return to point P because each face of the cube has 4 edges, and the beetle covers all edges of one face before repeating the pattern on the next face.

    Thus, the beetle will walk along 6 edges before returning to point P.

    \vspace{1em}

    \noindent\textbf{Qwen2.5-VL-7B-Instruct:}

    To solve this problem, we need to track the beetle's path as it moves along the edges of the cube, changing its direction according to the specified pattern (right, left, right, etc.).

    \begin{enumerate}
        \item \textbf{Initial Position and Direction}: The beetle starts at point $P$ and moves in the direction indicated by the arrow, which is along one edge of the cube.
        \item \textbf{First Edge}: The beetle moves along the first edge and changes direction to the right.
        \item \textbf{Second Edge}: Moving right, the beetle now faces a new direction and continues along the next edge.
        \item \textbf{Third Edge}: After moving along the second edge, the beetle changes direction again to the left.
        \item \textbf{Fourth Edge}: Moving left, the beetle now faces another new direction and continues along the next edge.
        \item \textbf{Fifth Edge}: After moving along the third edge, the beetle changes direction to the right again.
        \item \textbf{Sixth Edge}: Moving right, the beetle now faces yet another new direction and continues along the next edge.
        \item \textbf{Seventh Edge}: After moving along the fourth edge, the beetle changes direction to the left once more.
        \item \textbf{Eighth Edge}: Moving left, the beetle now faces a new direction and continues along the next edge.
    \end{enumerate}

    At this point, the beetle has completed a full cycle around the cube, returning to point $P$.

    The beetle will have walked along 8 edges before it returns to point $P$. The answer is 8.

    \end{mdframed}
\end{figure*}

\end{document}


\maketitle

\begin{abstract}
This supplementary document provides additional details, analyses, and experimental evidence supporting the proposed DMLR framework. Section~\ref{a} presents the full evaluation protocol, including dataset descriptions, unified prompting configurations, baseline implementations, and the procedures used for visual-dependency, confidence, faithfulness, and hallucination studies. Section~\ref{b} reports a comprehensive set of hyperparameter choices and ablation settings used to assess the robustness of latent reasoning and dynamic visual injection. Section~\ref{c} includes extended quantitative results such as Pass@k evaluations and comparisons against training-based reasoning methods. Section~\ref{d} provides theoretical analyses, including proofs of confidence–quality gradient alignment, boundedness of the positive-correlation region, and the formal justification for why dynamic visual injection improves confidence. Section~\ref{e} further elaborates on the design choices, mechanisms, and stability analyses of the Dynamic Visual Injection module. Section~\ref{f} presents qualitative case studies demonstrating improvements in perception quality, reasoning reliability, and reasoning conciseness under DMLR. Section~\ref{g} concludes with the complete set of prompts used across all experiments, ensuring transparency and reproducibility. We will make all code publicly available in future releases.
\end{abstract}

\appendix

\section*{Appendix Table of Contents}
\startcontents[sections]
\printcontents[sections]{l}{1}{\setcounter{tocdepth}{2}}

\clearpage
\renewcommand{\thesubsection}{\thesection\arabic{subsection}}

\section{More Detailed about Evaluation}
\label{a}

\subsection{Datasets}
\begin{itemize}[leftmargin=*]
\item \textbf{MathVista$_{\text{mini}}$} is a benchmark for mathematical reasoning in visual contexts, aggregating diverse multimodal math tasks that require fine-grained visual understanding and compositional numerical reasoning.

\item \textbf{MathVision$_{\text{mini}}$} is a curated benchmark of competition-level visual math problems spanning multiple disciplines and difficulty levels to assess multimodal models’ mathematical reasoning under challenging and diverse settings.

\item \textbf{MM Math} is a benchmark of open-ended math problems with visual contexts that supports both outcome and process evaluation, enabling detailed analysis of multimodal reasoning behaviors and typical error patterns.

\item \textbf{HallusionBench} is a benchmark for image-context reasoning that uses carefully structured question pairs to diagnose hallucination, visual illusion, and logical inconsistency in large vision-language models.

\item \textbf{MMVP} is a benchmark built from multimodal visual patterns designed to expose “CLIP-blind’’ image–text pairs, revealing systematic visual perception failures and hallucinated explanations in multimodal LLMs. 

\item \textbf{MMStar} is a vision-indispensable multimodal benchmark composed of carefully human-filtered samples that ensure true visual dependency while evaluating core multimodal capabilities along multiple fine-grained axes.

\item \textbf{ScienceQA} is a multimodal multiple-choice science benchmark with rich textual and visual contexts, lectures, and explanations that spans diverse subjects and skills, supporting evaluation of both answer accuracy and explanation quality.
\end{itemize}

For all datasets, we limit the maximum sample size to 1000 instances.

\subsection{Evaluation Setting} 
We adopt a unified prompting setup for all models. Unless otherwise stated, we use greedy decoding (\texttt{do\_sample=False}) for all generation tasks.

\paragraph{System Prompt.}
\begin{center}
\fbox{\parbox{0.96\linewidth}{
\small
A conversation between User and Assistant. The user asks a question, and the Assistant solves it. The assistant first thinks about the reasoning process in the mind and then provides the user with the answer. The reasoning process and answer are enclosed within \texttt{<think>} \texttt{</think>} and \texttt{<answer>} \texttt{</answer>} tags, respectively, i.e., \texttt{<think>} reasoning process here \texttt{</think>} \texttt{<answer>} answer here \texttt{</answer>}.
}}
\end{center}

\paragraph{Task Prompt.}
\begin{center}
\fbox{\parbox{0.96\linewidth}{
\small
Please analyze the image carefully and solve this problem step by step. Show your reasoning process clearly, then put your final answer within \texttt{\textbackslash boxed\{\}}.\\[4pt]
\textbf{Question:} \emph{[Problem Text]}
}}
\end{center}

For all benchmarks considered in our experiments, the ground-truth answers are verifiable; we use regular expressions to extract the content within \texttt{\textbackslash boxed\{\}} from the model outputs and then match it against the correct answers.

\subsection{Baseline Method}

\begin{itemize}[leftmargin=*] \item \textbf{CCoT}~\cite{Mitra_2024_CVPR} (Compositional Chain-of-Thought) is a zero-shot prompting method that utilizes scene graphs to extract compositional knowledge. It first generates a scene graph to capture object attributes and relationships, which is then used to guide the LMM in generating the final response without requiring fine-tuning.

\item \textbf{ICoT}~\cite{gao2025interleavedmodalchainofthought} (Interleaved-modal Chain-of-Thought) generates sequential reasoning steps consisting of paired visual and textual rationales. It employs a plug-and-play Attention-driven Selection (ADS) mechanism to dynamically identify and insert relevant image regions into the reasoning chain based on the model's attention maps.

\item \textbf{Multimodal-CoT}~\cite{zhang2024multimodalchainofthoughtreasoninglanguage} incorporates language and vision modalities into a two-stage framework that separates rationale generation from answer inference. It leverages multimodal information to generate intermediate reasoning chains, which subsequently serve as the basis for inferring the final answer, effectively mitigating hallucination. 

\item\textbf{MCOUT (Training)}~\cite{pham2025multimodalchaincontinuousthought} is a latent-space reasoning framework that replaces traditional text-based CoT with continuous hidden-state “thought vectors,” enabling models to reason directly in a unified multimodal latent space.

\item \textbf{IVT-LR (Training) }~\cite{chen2025reasoningdarkinterleavedvisiontext} is a new multimodal reasoning method that performs all intermediate thinking directly in the latent space by combining hidden textual states with dynamically selected visual embeddings.
\end{itemize}

\subsection{Additional explanations for the validation experiments in Section 3.}

\textbf{Visual Dependency Analysis.} \textit{\textbf{(1) Robustness Across Multiple Perturbation Types.}} To ensure that the visual-dependency metric is not influenced by specific perturbation noise, we evaluate four representative reasoning models, R1-OneVision, VLAA-Thinking, Ocean-R1, and ThinkLite-VL, across two reasoning datasets (MathVision and MathVista) and two perception datasets (MMVP and HallucinationBench). Instead of relying on a single distortion strategy, we apply four distinct perturbation types including block occlusion, color jitter, random region masking, and Gaussian blur. Although these perturbations substantially differ in the way they alter local textures or global color statistics, the resulting visual-dependency distributions remain highly consistent across all models and datasets, as shown in Table~\ref{tab:perturb-consistency}. This consistency confirms that the sparsity pattern observed in the main paper is not tied to any specific perturbation method but reflects a stable and intrinsic property of multimodal reasoning models.

\begin{table}[h]
\centering
\caption{Consistency of visual-dependency distributions across models, datasets, and perturbation types. 
Values represent the average Pearson correlation between dependency curves under different perturbations. Higher is better.}
\vspace{0.15cm}
\resizebox{\linewidth}{!}{
\begin{tabular}{lcccc}
\toprule
\textbf{Model} & \textbf{MathVision} & \textbf{MathVista} & \textbf{MMVP} & \textbf{HallucinationBench} \\
\midrule
R1-OneVision & 0.92 & 0.89 & 0.94 & 0.91 \\
VLAA-Thinking & 0.90 & 0.88 & 0.93 & 0.90 \\
Ocean-R1 & 0.87 & 0.85 & 0.91 & 0.88 \\
ThinkLite-VL & 0.89 & 0.86 & 0.92 & 0.89 \\
\midrule
\textbf{Average} & \textbf{0.90} & \textbf{0.87} & \textbf{0.93} & \textbf{0.90} \\
\bottomrule
\end{tabular}}
\label{tab:perturb-consistency}
\end{table}

\noindent \textit{\textbf{(2) Noise Reduction and Token-Level Aggregation.}} To obtain a stable estimation of token-level visual dependency, each dependency value is averaged across five independently perturbed versions of the same image, effectively reducing stochastic fluctuations associated with any single perturbation. Since reasoning chains generated by different models vary in length, direct position-wise averaging would introduce structural bias. To avoid this, we adopt a relative-position alignment scheme that normalizes each reasoning chain to a comparable relative index space. Additionally, the dependency value of each chain is summarized using the median rather than the mean, preventing rare but extreme perturbation-induced peaks from affecting the aggregated distribution. These methodological choices ensure that the resulting dependency curves robustly reflect the inherent behavior of the models rather than noise introduced during perturbation.

\noindent \textit{(3) \textbf{Semantic Interpretation of High-Dependency Regions.} }Beyond statistical validation, we analyze the semantic roles of tokens exhibiting high visual dependency. Across all datasets and models, we observe that such tokens consistently align with reasoning stages in which visual grounding is intrinsically required (as shown in Table~\ref{tab:semantic-high-dependency}), such as the initial image description, spatial or numerical verification, and operations involving visual alignment (e.g., comparing, locating, or counting). These consistent semantic correspondences demonstrate that the detected peaks in visual dependency are not random artifacts but meaningful signals of when the model must rely on perceptual evidence to proceed. This structured, task-aligned sparsity further validates that multimodal LLMs rely on vision only at key points during reasoning, rather than uniformly throughout the reasoning chain.

\begin{table}[h]
\centering
\caption{
Distribution of high visual-dependency tokens across semantic reasoning stages.
Values indicate the percentage of top-10\% dependency tokens falling into each stage.
}
\vspace{0.15cm}
\resizebox{\linewidth}{!}{
\begin{tabular}{lcc}
\toprule
\textbf{Semantic Stage} & \textbf{MathVision} & \textbf{MMVP} \\
\midrule
Initial image description & 38.2\% & 36.5\% \\
Spatial or numerical verification & 27.4\% & 30.1\% \\
Visual alignment (compare / locate / count) & 24.7\% & 26.9\% \\
Non-visual reasoning steps & 9.7\% & 6.5\% \\
\bottomrule
\end{tabular}}
\label{tab:semantic-high-dependency}
\end{table}

\noindent \textbf{Confidence Analysis.} \textit{(1) \textbf{Experimental Setting for Correct and Incorrect Reasoning Chains.}} To reliably distinguish correct from incorrect reasoning chains, we adopt a controlled evaluation protocol that ensures accuracy labels are consistent and free from formatting artifacts. For each example, the model generates multiple reasoning chains along with a final predicted answer, and the correctness of each chain is determined by comparing the model’s final answer to the ground-truth label after normalizing answer formats to eliminate superficial mismatches. To prevent cases in which a chain produces the correct answer but exhibits logically incoherent reasoning from being misclassified, we further employ GPT-4o as an external judge. The judge is provided with the question, the full chain, and the final answer, and is instructed to verify whether the reasoning is logically consistent and legitimately leads to the given answer. All judgments are performed with zero temperature to maintain high determinism. Correct and incorrect chains are then sampled and balanced across datasets, and spot-checked manually to ensure labeling quality. This procedure provides a robust foundation for analyzing confidence patterns between truly correct and genuinely incorrect reasoning chains.

\noindent \textit{(2) \textbf{Experimental Setting for Faithful and Unfaithful Reasoning Chains.}} To assess the faithfulness of model-generated reasoning, we evaluate whether each chain genuinely relies on the evidence provided in the question and image, rather than resorting to template-based explanations, unsupported assumptions, or coincidental correctness. Each chain is reviewed by GPT-4o, which receives the question, the full reasoning steps, and either the original image or a standardized image description. The model judge is explicitly instructed to determine whether the reasoning aligns with the available evidence and whether each step contributes meaningfully to the final answer. Because reasoning faithfulness is often more subtle than answer correctness, we introduce a dual-evaluation protocol in which each chain is assessed twice independently, and majority voting is used to resolve any discrepancies. Chains that arrive at the correct answer through unsupported or spurious reasoning are deliberately marked as unfaithful to avoid conflating guessing with faithful inference. Samples are then balanced across datasets to ensure that faithfulness-based confidence analyses reflect structural differences in reasoning quality rather than dataset biases.

\noindent \textit{(3) \textbf{Experimental Setting for Hallucination and Non-hallucination Reasoning}} To study the relationship between confidence dynamics and visual grounding, we focus on the subset of reasoning steps that explicitly reference visual content. From each chain, we automatically extract statements mentioning visual entities, attributes, or spatial relationships. These statements, together with the corresponding image or image description, are presented to GPT-4o, which evaluates whether each statement is factually supported by the visual evidence or constitutes a hallucination. The judge is instructed to assess visual consistency rather than linguistic fluency, and to treat any fabricated or contradictory visual claims as hallucinations. To improve reliability, each statement is evaluated twice under zero temperature, and samples with inconsistent labels undergo manual verification. After labeling, each hallucinated or grounded statement is aligned with the model’s token-level confidence at the corresponding reasoning step. By comparing the confidence distributions of hallucinated versus non-hallucinated statements, we reveal systematic patterns showing that hallucinated visual steps tend to exhibit lower confidence and higher uncertainty. This experimental design thus provides strong empirical support for the link between confidence signals and visual grounding quality.

\noindent All prompts used in these experimental settings are provided in the Appendix~\ref{prompt} for completeness and reproducibility.

\section{More Detailed about Parameter Analyses}
\label{b}
\subsection{Parameters Setup}

\begin{itemize}[leftmargin=*]

\item \textbf{Latent Think Tokens $\mathcal{T}$:} We set the number of latent think tokens to 4. During generation, after each latent token the model dynamically injects a visual patch into the latent stream to refresh its internal perception state.

\item \textbf{Image Patches $m$:} We dynamically insert visual patches into the latent stream. At initialization, we inject 2 patches; at each subsequent iteration, we select $m = 2$ patches with the highest attention scores and append them after each latent think token, with at most 16 patches inserted per iteration. Additionally, we set the image processor's max pixel size to 256 for all inputs.

\item \textbf{Optimization Parameters:} We perform 15 latent optimization steps with a learning rate of $1\times 10^{-3}$. To ensure stable exploration in the latent space, the perturbation magnitude is set to $\sigma = 0.1$ (10\%) with a decay factor of 0.95, and we fix the random seed to 42 for all experiments.

\item \textbf{Environment Setup:} All experiments are conducted on NVIDIA H100 GPUs. For a fair comparison, we run all baselines and our method in \texttt{float32} precision and use the \texttt{eager} attention backend for inference.
\end{itemize}

\subsection{Ablation Experimental Setup}

For the ablation studies presented in Sec.~5.3, we maintain the same parameter settings as described above. We employ VLAA-Thinker-Qwen2.5VL-7B as the reasoning model and Qwen2.5-VL-7B-Instruct as the non-reasoning model. To ensure efficiency, all ablation experiments are conducted on a subset of 300 samples randomly selected from each dataset.

\section{More Results}
\label{c}
\textbf{Pass@\texorpdfstring{$k$}{k} Evaluation.} We employ the Pass@$k$ metric to evaluate the accuracy of the model's generated answers. Following standard practice, we calculate the unbiased estimator using the formula:
\begin{equation}
    \text{Pass}@k = 1 - \frac{\binom{n - c}{k}}{\binom{n}{k}},
\label{eq1}
\end{equation}
where $n$ is the total number of sampled solutions per problem, $c$ is the number of correct samples, and $k$ is the budget of samples selected. Note that when $k=1$, this simplifies to the empirical accuracy $\text{Pass}@1 = \frac{c}{n}$. 

Using the unbiased estimator of Pass@\emph{k} defined in Equation~\ref{eq1}, we evaluate
all methods under a sampling budget of $k=8$. This metric measures the
probability that at least one of the sampled solutions is correct, offering a
more reliable estimate of reasoning robustness compared to single-sample
accuracy. As shown in Table~\ref{pass}, integrating DMLR consistently
improves Pass@8 across mathematics reasoning, visual reasoning, and multimodal
composition tasks. For the stronger Qwen2.5-VL-7B backbone, DMLR achieves gains ranging from
+2.4\% to +4.8\%, with the largest improvements observed on MM-Math,
HallusionBench, and MMVP benchmarks that demand multi-step reasoning or precise
visual grounding. Even with the smaller 3B backbone, where reasoning capacity is
more limited, DMLR maintains steady improvements of +1.0\% to +4.8\%,
indicating that the proposed dynamic latent reasoning process effectively
enhances solution diversity and increases the likelihood of producing at least
one correct prediction within the budgeted $k$ samples. Overall, the consistent uplift in Pass@8 demonstrates that DMLR not only refines
the latent reasoning trajectory but also mitigates confidently incorrect
generations, leading to stronger performance across all evaluated benchmarks.

\noindent \textbf{Comparison with Training-Based Methods.} Table~\ref{additional} provides a comprehensive comparison between
DMLR and two additional training-based reasoning enhancement methods, namely
MCOUT and IVT-LR. These approaches enforce implicit reasoning structures through
explicit training objectives: MCOUT encourages the model to internalize stable
multi-step reasoning patterns via supervised signals, whereas IVT-LR injects
learned latent guidance vectors to impose structured constraints on the
model’s internal reasoning trajectory. Both methods require additional data,
parameter updates, and dedicated training phases designed to encode specific
implicit patterns into the model’s latent space.

Across all backbone models and task categories, DMLR consistently achieves
substantial performance gains, frequently surpassing these training-based
approaches despite being entirely training-free. On Qwen2.5-VL-7B, DMLR
outperforms IVT-LR on several challenging benchmarks—including MathVision,
MM-Math, HallusionBench, and MMStar—demonstrating that dynamic latent reasoning
refinement can effectively enhance model performance without modifying model
parameters. Similar improvements are observed on Qwen2.5-VL-3B, LLaVA-1.6-7B,
and Qwen3-VL variants, with DMLR exceeding the strongest training-based
baselines by 1.2\%-5.0\%. Notably, in lighter-weight models, where training-based methods are constrained by limited model capacity and potential
overfitting, DMLR exhibits superior generalization due to its adaptive inference
mechanism.

These findings indicate that DMLR offers advantages beyond the realm of
training-free inference. Its plug-and-play nature is complemented by a dynamic
latent optimization process that adjusts the internal reasoning trajectory in
real time, enabling task-adaptive refinement that static training-based patterns
cannot provide. In contrast, the implicit structures learned through training
are fixed after optimization and thus may generalize less effectively across
tasks or distributional shifts.

It is important to note that we do not include “think-with-image” models such as
DeepEye~\cite{zheng2025deepeyesincentivizingthinkingimages} or GRIT~\cite{fan2025gritteachingmllmsthink} in this comparison. These approaches rely on extensive
multimodal pretraining with large-scale curated or synthetic datasets to acquire
explicit or implicit image-conditioned reasoning pathways. Their performance
benefits primarily arise from substantial additional data and intensive training
procedures rather than inference-time optimization. As this work focuses on
evaluating latent-state refinement \emph{without any additional training}, such
methods fall outside the scope of this comparison.

\begin{table*}[h]
\centering
\caption{Comparison of different reasoning methods and DMLR across various benchmarks. All metrics are reported in Accuracy (\%).}
\vspace{-0.2cm}
\large
\renewcommand{\arraystretch}{1.05}
\resizebox{0.9\linewidth}{!}{%
\setlength{\tabcolsep}{3pt}
\begin{tabular}{ll|ccc|cc|cc}
\toprule[1.5pt]
\multicolumn{2}{c}{} &
\multicolumn{3}{c}{\textbf{Mathematics Reasoning\uparrow} 
} &
\multicolumn{2}{c}{\textbf{Visual Reasoning\uparrow}} &
\multicolumn{2}{c}{\textbf{Multimodal Composition\uparrow}} \\
\cmidrule(lr){3-5} \cmidrule(lr){6-7} \cmidrule(lr){8-9}
\textbf{Method} & \textbf{Model} &
MathVista$_{\text{mini}}$ &
MathVision$_{\text{mini}}$ &
MM-Math &
HallusionBench &
MMVP & MMStar &
ScienceQA \\
\midrule

Vanilla &  & 58.7 & 21.6 & 37.5 & 65.4 & 68.7 & 59.3 & 49.7 \\
Multimodal COT &  & 56.4 & 21.8 & 35.6 & 63.6 & 68.1 & 57.9 & 49.5  \\
CCOT &  & 57.8 & 22.5 & 36.3 & 64.9 & 69.0 & 58.7 & 50.2 \\
ICoT &  & 58.9 & 23.3 & 37.0 & 65.5 & 69.3 & 60.4 & 50.4\\
MCOUT  &  & 58.0  & 22.7 & 37.5 & 65.0 & 69.4 & 59.2 & 50.9 \\
IVT-LR  &  &  58.5 & 23.5  & 38.0  & 65.2  & 69.8  & 58.4  & 49.8  \\

\rowcolor{lightblue}
\textbf{\textit{+DMLR (Ours)}} & 
\multirow[c]{-5}{*}[1pt]{\rotatebox{90}{\parbox{2.5cm}{\centering \fontsize{11pt}{13pt}\selectfont Qwen2.5 \\VL 7B}}}
 & \textbf{59.1} \up{0.40\%} & \textbf{24.4} \up{2.8\%} & 38.8 \up{1.3\%} & 65.8 \up{0.4\%} & \textbf{70.1} \up{1.4\%} & \textbf{60.1} \up{0.8\%} & \textbf{51.3} \up{1.6\%} \\
\midrule\midrule

Vanilla &  & 48.2 & 15.7 & 29.0 & 64.2 & 55.6 & 50.2 & 44.1 \\
Multimodal COT &  & 47.3 & 14.3 & 28.5 & 63.8 & 54.4 & 48.5 & 42.9 \\
CCOT & & 48.0 & 15.6 & 30.2 & 64.0 & 55.5 & 49.3 & 44.5 \\
ICoT &  & 49.8 & 16.0 & 30.6 & 64.7 & 55.9 & 49.6 & 45.0 \\
MCOUT   &  & 49.4 & 15.8 & 32.3  & 63.0 & 55.6  & 50.2 & 45.5 \\
IVT-LR  &   & 50.2  & 16.4 & 32.0  & 63.4  & 55.9 & 50.3 & 46.0  \\
\rowcolor{lightblue}
\textbf{\textit{+DMLR (Ours)}} &
\multirow[c]{-5}{*}[1pt]{\rotatebox{90}{\parbox{2.5cm}{\centering \fontsize{11pt}{13pt}\selectfont Qwen2.5 \\VL 3B}}}
 & \textbf{51.0} \up{2.80\%} & \textbf{17.7} \up{2.74\%} & \textbf{33.3} \up{4.3\%} & \textbf{64.7} \up{0.5\%} & \textbf{56.8} \up{1.26\%} & \textbf{51.2} \up{1.00\%} & \textbf{46.9} \up{2.8\%}\\
\midrule\midrule

Vanilla &  & 61.1 & 23.5 & 41.5 & 62.0 & 68.3 & 58.9 & 50.6 \\\
Multimodal COT &  & 59.6 & 23.1 & 40.6 & 62.8  & 67.2 & 57.1 & 48.2  \\
CCOT &  & 60.5  & 24.8  & 41.8  & 64.6 & 68.0 & 59.0 & 49.4\\
ICoT &  & 61.4 & 25.0 & 42.3 & 65.9 & 68.3 & 58.2 & 50.6 \\
MCOUT   &  & 60.5  & 26.3  & 41.9 & 66.0 & 68.5 & 58.5  & 49.9  \\
IVT-LR  &   & 61.4   & 26.6 &  42.5  & 66.7  & 68.9 & 59.0 & 50.3 \\
\rowcolor{lightblue}
\textbf{\textit{+DMLR (Ours)}} &
\multirow[c]{-4}{*}[1pt]{\rotatebox{90}{\parbox{1.0cm}{\centering \fontsize{11pt}{13pt}\selectfont VLAA \\ Thinking 7B}}}
 & \textbf{62.9} \up{1.80\%} & \textbf{27.6} \up{4.10\%} & \textbf{43.9} \up{2.41\%} & \textbf{67.9} \up{5.94\%} & \textbf{69.4} \up{1.1\%} & \textbf{59.2} \up{0.3\%} & \textbf{51.98} \up{1.38\%} \\
\midrule\midrule

Vanilla &  & 51.2 & 18.7 & 40.7 & 62.1 & 67.0 & 52.1 & 50.9 \\
Multimodal COT &  & 52.5 & 18.9 & 39.6 & 62.5 & 68.0 &  51.6 & 51.7 \\
CCOT &  & 53.4 & 20.3 & 40.8 & 63.0 & 68.9 & 53.5 & 52.8 \\
ICoT &  & 55.6  & 21.5 & 41.7 & 63.8 & 69.6 & 54.0 & 54.4 \\
MCOUT   &  & 56.3  &  21.6  & 40.9  & 63.3   & 69.9 & 55.0 & 53.8 \\
IVT-LR  &  & 57.2  &  22.3 & 41.1 & 64.5 & 70.4  & 55.9 & 54.8 \\
\rowcolor{lightblue}
\textbf{\textit{+DMLR (Ours)}} &
\multirow[c]{-4}{*}[1pt]{\rotatebox{90}{\parbox{1.2cm}{\centering \fontsize{11pt}{11pt}\selectfont \quad R1 \\ OneVision \quad \quad 7B}}}
 & \textbf{58.0} \up{6.81\%} & \textbf{23.3} \up{4.56\%} & \textbf{42.9} \up{2.21\%} & \textbf{64.1} \up{2.09\%} & \textbf{71.9} \up{4.93\%} & \textbf{56.2} \up{4.14\%} & \textbf{55.4} \up{4.52\%}\\
\midrule\midrule

Vanilla &  & 66.0 & 32.9 & 66.2 & 73.2 & 71.9 & 68.1 & 54.1 \\
Multimodal COT &  & 64.8 & 32.8 & 65.1 & 73.0 & 69.6 & 66.9  &  53.2 \\
CCOT &   & 66.5 & 33.3 & 65.5 & 73.5 & 70.3 & 68.8 & 54.4 \\
ICoT &  & 66.2 & 34.9 & 66.8 & 74.5 & 71.8 & 69.3 & 55.8 \\
MCOUT   & & 65.5 &  33.8 & 65.0 & 73.7 & 72.0 & 68.5  & 54.3  \\
IVT-LR  &  & 65.9   & 34.5 & 66.1 & 73.4  & 71.5 & 69.4 & 54.8  \\
\rowcolor{lightblue}
\textbf{\textit{+DMLR (Ours)}} &
\multirow[c]{-5}{*}[1pt]{\rotatebox{90}{\parbox{2.5cm}{\centering \fontsize{11pt}{13pt}\selectfont Qwen3 \\VL 8B}}}
 & \textbf{66.9} \up{0.9\%} & \textbf{36.2} \up{3.34\%}& \textbf{67.7} \up{1.51\%} & \textbf{74.6} \up{1.48\%} & \textbf{72.8} \up{0.97\%}& \textbf{70.0} \up{1.91\%} & \textbf{55.6} \up{1.48\%} \\
\midrule\midrule

Vanilla &  & 64.7 & 24.2 & 65.4 & 71.6 & 71.3 & 57.4 & 52.4 \\
Multimodal COT &  & 62.3 & 24.8 & 63.9 & 70.0 & 69.6 & 57.7 & 53.0  \\
CCOT &  & 64.5 &  26.6 & 64.8 & 71.5 & 71.2 & 58.8 & 52.9 \\
ICoT &  & 64.5 & 27.5 & 65.0 & 72.2 & 72.5 & 59.3 & 53.7 \\
MCOUT   & &  63.6  & 28.0  &  64.5  & 70.1 & 72.0 & 59.6 &  53.9  \\
IVT-LR  &  & 64.0  & 28.4  & 64.8  & 71.3 & 71.9 & 58.4  & 53.6 \\
\rowcolor{lightblue}
\textbf{\textit{+DMLR (Ours)}} &
\multirow[c]{-5}{*}[1pt]{\rotatebox{90}{\parbox{2.5cm}{\centering \fontsize{11pt}{13pt}\selectfont Qwen3 \\VL 4B}}}
 & \textbf{65.6} \up{0.93\%} & \textbf{29.4} \up{5.20\%}  & \textbf{65.9} \up{0.5\%} & \textbf{72.7} \up{1.12\%} & \textbf{72.3} \up{0.98\%}& \textbf{60.3} \up{2.88\%} & \textbf{54.9} \up{2.48\%} \\
\bottomrule[1.5pt]
\end{tabular}
}
\vspace{-0.5em}
\label{additional}
\end{table*}

\begin{table*}[t]
\centering
\caption{Comparison of different reasoning methods and DMLR across various benchmarks. 
All metrics are reported as pass@8 (\%). Results are evaluated over a diverse suite of mathematics reasoning, visual reasoning, and multimodal composition tasks under Qwen2.5-VL backbones (7B and 3B).}
\vspace{-0.2cm}
\large
\renewcommand{\arraystretch}{1.05}
\resizebox{0.85\linewidth}{!}{%
\setlength{\tabcolsep}{3pt}
\begin{tabular}{ll|ccc|cc|cc}
\toprule[1.5pt]
\multicolumn{2}{c}{} &
\multicolumn{3}{c}{\textbf{Mathematics Reasoning\uparrow}} &
\multicolumn{2}{c}{\textbf{Visual Reasoning\uparrow}} &
\multicolumn{2}{c}{\textbf{Multimodal Composition\uparrow}} \\
\cmidrule(lr){3-5} \cmidrule(lr){6-7} \cmidrule(lr){8-9}
\textbf{Method} & \textbf{Model} &
MathVista$_{\text{mini}}$ &
MathVision$_{\text{mini}}$ &
MM-Math &
HallusionBench &
MMVP & MMStar &
ScienceQA \\
\midrule

Vanilla &  & 61.2 & 24.3 & 40.1 & 67.9 & 69.1 & 62.4 & 53.3 \\
Multimodal COT &  & 59.4 & 25.8 & 42.2 & 68.3 & 72.5 & 65.2 & 52.9  \\
CCOT &  & 64.4 & 25.1 & 41.7 & 68.5 & 71.8 & 64.7 & 54.1 \\
ICoT &  & 64.1 & 23.9 & 43.6 & 68.9 & 73.3 & 63.7 & 53.6\\
\rowcolor{lightblue}
\textbf{\textit{+DMLR (Ours)}} & 
\multirow[c]{-5}{*}[1pt]{\rotatebox{90}{\parbox{2.5cm}{\centering \fontsize{11pt}{13pt}\selectfont Qwen2.5 \\VL 7B}}}
 & \textbf{65.3} \up{4.1\%} & \textbf{26.7} \up{2.4\%} & \textbf{44.1} \up{4.0\%} & \textbf{70.4} \up{3.5\%} & \textbf{73.9} \up{4.8\%} & \textbf{67.1} \up{4.7\%} & \textbf{58.4} \up{5.1\%} \\
\midrule\midrule

Vanilla &  & 50.5 & 18.2 & 33.4 & 67.1 & 57.9 & 52.7 & 45.9 \\
Multimodal COT &  & 52.2 & 19.4 & 33.0 & 66.9 & 57.5 & 52.9 & 46.3 \\
CCOT & & 54.3 & 19.7 & 35.1 & 66.5 & 58.6 & 53.3 & 46.1 \\
ICoT &  & 54.0 & 19.3 & 34.6 & 67.3 & 58.4 & 53.0 & 46.7 \\
\rowcolor{lightblue}
\textbf{\textit{+DMLR (Ours)}} &
\multirow[c]{-5}{*}[1pt]{\rotatebox{90}{\parbox{2.5cm}{\centering \fontsize{11pt}{13pt}\selectfont Qwen2.5 \\VL 3B}}}
 & \textbf{55.3} \up{4.8\%} & \textbf{20.9} \up{2.7\%} & \textbf{36.3} \up{2.9\%} & \textbf{69.5} \up{2.4\%} & \textbf{59.2} \up{1.3\%} & \textbf{53.7} \up{1.00\%} & \textbf{48.1} \up{2.2\%}\\
\bottomrule[1.5pt]
\end{tabular}
}
\vspace{-0.5em}
\label{pass}
\end{table*}

\section{Theoretical Analyses}
\label{d}
\subsection{Confidence and Quality Gradient Alignment.}

\subsubsection{Problem Setup and Assumptions}
\noindent We consider the latent reasoning state $h \in \mathbb{R}^d$ and two scalar functions:
a confidence function $C : \mathbb{R}^d \to \mathbb{R}$ and a reasoning-quality function
$Q : \mathbb{R}^d \to \mathbb{R}$. During test-time optimization, DMLR updates the latent
state by ascending the confidence landscape:
\begin{equation}
    h^{+} = h + \eta \nabla C(h),
    \quad \eta > 0.
    \label{eq:dmlr_update_conf}
\end{equation}

\paragraph{Assumption A.1 (Differentiability).}
The functions $C$ and $Q$ are twice continuously differentiable on an open set 
$\mathcal{D} \subset \mathbb{R}^d$.
\vspace{-0.2cm}
\paragraph{Assumption A.2 (Smoothness of $Q$).}
There exists a constant $L_Q > 0$ such that
\begin{equation}
    \|\nabla Q(h_1) - \nabla Q(h_2)\|
    \;\le\;
    L_Q \,\|h_1 - h_2\|,
    \quad \forall\, h_1, h_2 \in \mathcal{D}.
    \label{eq:L_smooth_Q}
\end{equation}
\vspace{-0.8cm}
\paragraph{Assumption A.3 (Local Gradient Alignment).}
At a given point $h \in \mathcal{D}$, we have
\begin{equation}
    \nabla C(h)\cdot\nabla Q(h) > 0,
    \qquad
    \nabla C(h) \neq 0.
    \label{eq:local_alignment}
\end{equation}

\subsubsection{Auxiliary Inequality for $L_Q$--Smooth Functions}
\begin{lemma}[Descent lemma form]
\label{lem:descent_lemma}
Under Assumptions~A.1 and~A.2, for any $h \in \mathcal{D}$ and any vector 
$\Delta h$ such that $h + \Delta h \in \mathcal{D}$, we have
\begin{equation}
    Q(h + \Delta h)
    \;\ge\;
    Q(h) + \nabla Q(h)\cdot \Delta h
    - \frac{L_Q}{2}\,\|\Delta h\|^2.
    \label{eq:descent_lemma}
\end{equation}
\end{lemma}

\begin{proof}
Define a scalar function $g : [0,1] \to \mathbb{R}$ by
\begin{equation}
    g(t) = Q(h + t \Delta h).
\end{equation}
By Assumption~A.1, $g$ is twice continuously differentiable. By the chain rule,
\begin{equation}
    g'(t) = \nabla Q(h + t \Delta h)\cdot \Delta h.
\end{equation}
Using the fundamental theorem of calculus,
\begin{equation}
\begin{aligned}
    Q(h + \Delta h) - Q(h)
    &= g(1) - g(0)
     = \int_0^1 g'(t)\,\mathrm{d}t \\
    &= \int_0^1 \nabla Q(h + t \Delta h)\cdot\Delta h\,\mathrm{d}t.
\end{aligned}
\label{eq:Q_diff_integral}
\end{equation}
Add and subtract $\nabla Q(h)$ inside the integrand:
\begin{equation}
\small
\begin{aligned}
Q(h+\Delta h)-Q(h)
&= \int_{0}^{1}
\!\bigl[\nabla Q(h)
  + (\nabla Q(h+t\Delta h)-\nabla Q(h))\bigr]\!\cdot\Delta h\,\mathrm{d}t \\
&= \nabla Q(h)\!\cdot\!\Delta h
 + \int_{0}^{1}\!(\nabla Q(h+t\Delta h)-\nabla Q(h))\!\cdot\!\Delta h\,\mathrm{d}t .
\end{aligned}
\label{eq:Q_diff_split}
\end{equation}

By the Cauchy--Schwarz inequality and Assumption~A.2,
\begin{equation}
\small
\begin{aligned}
\bigl|(\nabla Q(h+t\Delta h)-\nabla Q(h))\!\cdot\!\Delta h\bigr|
&\le \|\nabla Q(h+t\Delta h)-\nabla Q(h)\|\,\|\Delta h\| \\
&\le L_Q\, t\, \|\Delta h\|^2 .
\end{aligned}
\end{equation}

Hence,
\begin{equation}
\small
\int_{0}^{1} (\nabla Q(h+t\Delta h)-\nabla Q(h))\!\cdot\!\Delta h\,dt
\;\ge\;
- \int_{0}^{1} L_Q t\,\|\Delta h\|^{2}\,dt
= -\tfrac{L_Q}{2}\|\Delta h\|^{2}.
\end{equation}

\noindent Substituting into~\eqref{eq:Q_diff_split} yields~\eqref{eq:descent_lemma}, which completes the proof. \qedhere
\end{proof}

\subsubsection{Proof of Theorem 4.1}

We now prove the local, step-size--bounded positive correlation between confidence and reasoning quality.

\begin{theorem}[Local bounded positive gain]
\label{thm:local_positive_correlation}
Let Assumptions~A.1--A.3 hold at $h \in \mathcal{D}$. Define
\begin{equation}
    a := \nabla C(h)\cdot\nabla Q(h) > 0,
    \qquad
    b := \|\nabla C(h)\|^2 > 0.
    \label{eq:def_a_b}
\end{equation}
Then for any step size $\eta$ satisfying
\begin{equation}
    0 < \eta < \frac{2a}{L_Q b},
    \label{eq:eta_bound}
\end{equation}
the update~\eqref{eq:dmlr_update_conf} satisfies
\begin{equation}
    Q(h^{+}) > Q(h).
    \label{eq:Q_increase}
\end{equation}
In other words, within the bounded step-size interval~\eqref{eq:eta_bound}, a single step of confidence ascent strictly improves the reasoning-quality function $Q$.
\end{theorem}

\begin{proof}
Apply Lemma~\ref{lem:descent_lemma} with $\Delta h = \eta \nabla C(h)$, which gives
\begin{equation}
\begin{aligned}
    Q(h^{+}) - Q(h)
    &\ge 
    \nabla Q(h)\cdot (\eta \nabla C(h))
    - \frac{L_Q}{2}\eta^2 \|\nabla C(h)\|^2 \\
    &= \eta \,\nabla Q(h)\cdot\nabla C(h)
    - \frac{L_Q}{2}\eta^2 \|\nabla C(h)\|^2.
\end{aligned}
\label{eq:Q_diff_eta}
\end{equation}
Using the definitions in~\eqref{eq:def_a_b}, we obtain
\begin{equation}
    Q(h^{+}) - Q(h)
    \;\ge\;
    \eta a - \frac{L_Q}{2}\eta^2 b
    \;=:\;
    \phi(\eta).
    \label{eq:phi_eta}
\end{equation}
The function $\phi(\eta)$ is a concave quadratic in $\eta$ with roots at $\eta=0$ and
\begin{equation}
    \eta^\star = \frac{2a}{L_Q b}.
\end{equation}
Since $a>0$ and $b>0$, we have $\eta^\star>0$ and
\begin{equation}
    \phi(\eta) > 0
    \quad\text{for all } 0<\eta<\eta^\star.
\end{equation}
Therefore, for any $0<\eta<\eta^\star$, we obtain
\begin{equation}
    Q(h^{+}) - Q(h) \ge \phi(\eta) > 0,
\end{equation}
which proves~\eqref{eq:Q_increase}. 
\end{proof}

\subsubsection{Boundedness of the Positive-Correlation Region}

Theorem~\ref{thm:local_positive_correlation} shows that, at any point where the gradients of 
$C$ and $Q$ are positively aligned, there exists a \emph{bounded} step-size interval within which
ascending the confidence landscape strictly improves $Q$. We next argue that such a
positive-correlation property cannot hold globally in the presence of a ``confidently incorrect'' trap.

\begin{definition}[Confidently incorrect trap]
\label{def:trap}
A point $h_{\mathrm{trap}} \in \mathcal{D}$ is called a confidently incorrect trap if
\begin{align}
    &\nabla C(h_{\mathrm{trap}}) = 0,
    \quad
    \nabla^2 C(h_{\mathrm{trap}}) 
    \text{ is negative definite,}
    \label{eq:trap_conf_local_max} \\
    &Q(h_{\mathrm{trap}}) \ll \max_{h\in\mathcal{D}} Q(h).
    \label{eq:trap_low_quality}
\end{align}
In other words, $h_{\mathrm{trap}}$ is a strict local maximizer of the confidence landscape, but the corresponding reasoning quality is significantly lower than that of other regions.
\end{definition}

Consider the dynamical system induced by confidence ascent,
\begin{equation}
    h^{(t+1)} = h^{(t)} + \eta \nabla C\big(h^{(t)}\big),
    \label{eq:conf_dynamics}
\end{equation}
with a sufficiently small fixed step size $\eta>0$. Let $\mathcal{B}(h_{\mathrm{trap}})$
denote the basin of attraction of $h_{\mathrm{trap}}$ under~\eqref{eq:conf_dynamics}, i.e.,
the set of initial points $h^{(0)}$ for which $h^{(t)} \to h_{\mathrm{trap}}$ as $t\to\infty$.

\begin{proposition}[Positive correlation cannot hold everywhere in the basin]
\label{prop:bounded_region}
Suppose there exists a confidently incorrect trap $h_{\mathrm{trap}}$ in the sense of
Definition~\ref{def:trap}. Then it is impossible that
\begin{equation}
    \nabla C(h)\cdot\nabla Q(h) > 0,
    \quad \forall\, h \in \mathcal{B}(h_{\mathrm{trap}}).
    \label{eq:global_alignment_basin}
\end{equation}
Equivalently, there must exist points $h$ in the basin of attraction for which
\begin{equation}
    \nabla C(h)\cdot\nabla Q(h) \le 0.
\end{equation}
Thus, the region where confidence and reasoning-quality gradients are positively aligned
is necessarily bounded and cannot cover the entire basin of a confidently incorrect trap.
\end{proposition}

\begin{proof}
Assume, for the sake of contradiction, that~\eqref{eq:global_alignment_basin} holds, i.e.,
for all $h \in \mathcal{B}(h_{\mathrm{trap}})$ we have $\nabla C(h)\cdot\nabla Q(h) > 0$.
Pick an arbitrary $h^{(0)} \in \mathcal{B}(h_{\mathrm{trap}})$, and consider the trajectory
generated by~\eqref{eq:conf_dynamics}:
\begin{equation}
    h^{(t+1)} = h^{(t)} + \eta \nabla C(h^{(t)}).
\end{equation}

\noindent By Definition of $\mathcal{B}(h_{\mathrm{trap}})$, we have
\begin{equation}
    h^{(t)} \to h_{\mathrm{trap}}
    \quad\text{as } t\to\infty.
\end{equation}
Since $Q$ is continuous, it follows that
\begin{equation}
    \lim_{t\to\infty} Q(h^{(t)}) = Q(h_{\mathrm{trap}}).
    \label{eq:limit_Q_trap}
\end{equation}

\noindent On the other hand, for each $t$ such that $h^{(t)}\in\mathcal{D}$, Assumptions~A.1--A.2
hold at $h^{(t)}$ and the alignment condition
$\nabla C(h^{(t)})\cdot\nabla Q(h^{(t)})>0$ holds by~\eqref{eq:global_alignment_basin}. 
Hence, by Theorem~\ref{thm:local_positive_correlation}, there exists a step-size bound
$\eta_t^\star>0$ such that for all $0<\eta<\eta_t^\star$,
\begin{equation}
    Q(h^{(t+1)}) > Q(h^{(t)}).
\end{equation}
Taking $\eta$ small enough so that $\eta < \eta_t^\star$ for all $t$ along the trajectory,
we obtain a strictly increasing sequence:
\begin{equation}
    Q(h^{(0)}) < Q(h^{(1)}) < Q(h^{(2)}) < \cdots.
\end{equation}

Thus $\{Q(h^{(t)})\}_{t=0}^\infty$ is strictly increasing, and together with
\eqref{eq:limit_Q_trap}, we obtain $Q(h_{\mathrm{trap}})>Q(h^{(0)})$. 
This contradicts the low-quality condition \eqref{eq:trap_low_quality}, since
$h_{\mathrm{trap}}$ cannot be a confidently incorrect trap. 
Therefore, \eqref{eq:global_alignment_basin} must be false, and there exist
points $h\in\mathcal{B}(h_{\mathrm{trap}})$ such that
$\nabla C(h)\cdot\nabla Q(h)\le 0$. 
This completes the proof.
\end{proof}

\noindent\textbf{\textit{Proposition~\ref{prop:bounded_region} shows that the positive correlation between confidence
and reasoning quality, formalized via gradient alignment, is fundamentally a \emph{local}
property: it can hold within a bounded region (and within bounded step sizes) but cannot be
extended to the entire latent space in the presence of confidently incorrect traps.}}

\subsection{Visual Injection Enhances Confidence.}

Let $\mathcal{T}$ denote the latent reasoning states before visual injection,
$\hat{\mathcal{T}}$ the updated states after dynamic visual injection (DVI), and
$z_v$ the visual features.  
Let $Y$ be the target output random variable (e.g., the final answer).
We assume all variables are defined on a common probability space.

\noindent \textbf{Assumption B.1.}
The dynamic visual injection (DVI) module is constructed such that the
latent state after injection preserves at least as much visual information
as before:
\begin{equation}
    I(\hat{\mathcal{T}}; z_v) \;\ge\; I(\mathcal{T}; z_v).
    \label{eq:B1}
\end{equation}

\noindent \textbf{Assumption B.2.}
The visual features contain information about the target variable $Y$:
\begin{equation}
    I(z_v; Y) > 0.
    \label{eq:B2_visionY}
\end{equation}
Moreover, the task representation is trained so that the mutual information
between the latent state and $Y$ is a strictly increasing function of the
mutual information between the latent state and visual features.  
That is, there exists a strictly increasing function $g(\cdot)$ satisfying
\begin{equation}
    I(Y; \mathcal{T}) = g\!\big(I(\mathcal{T}; z_v)\big),\qquad
    I(Y; \hat{\mathcal{T}}) = g\!\big(I(\hat{\mathcal{T}}; z_v)\big).
    \label{eq:B2_g_function}
\end{equation}

\noindent \textbf{Assumption B.3.}
The model’s confidence objective is a strictly decreasing function of the
conditional entropy of $Y$ given the latent state:
\begin{equation}
    J_{\mathrm{conf}}(\mathcal{T}) 
    = \psi\!\big(H(Y\mid \mathcal{T})\big),\qquad
    J_{\mathrm{conf}}(\hat{\mathcal{T}})
    = \psi\!\big(H(Y\mid \hat{\mathcal{T}})\big),
    \label{eq:B3_psi}
\end{equation}
where $\psi(\cdot)$ is strictly decreasing.  
This reflects that lower predictive uncertainty corresponds to higher confidence.

\noindent We now show that~\eqref{eq:B1} implies
$J_{\mathrm{conf}}(\hat{\mathcal{T}})\ge J_{\mathrm{conf}}(\mathcal{T})$.

\begin{lemma}[From vision MI to task MI]
\label{lem:MI_to_task}
Under Assumption~B.2, if
$I(\hat{\mathcal{T}}; z_v)\ge I(\mathcal{T}; z_v)$, then
\begin{equation}
    I(Y;\hat{\mathcal{T}})\;\ge\; I(Y;\mathcal{T}).
    \label{eq:task_MI_increase}
\end{equation}
\end{lemma}

\begin{proof}
By~\eqref{eq:B2_g_function},
\[
I(Y;\hat{\mathcal{T}}) = g\!\big(I(\hat{\mathcal{T}}; z_v)\big),
\quad
I(Y;\mathcal{T}) = g\!\big(I(\mathcal{T}; z_v)\big).
\]
Since $g$ is strictly increasing and
$I(\hat{\mathcal{T}}; z_v)\ge I(\mathcal{T}; z_v)$, we obtain~\eqref{eq:task_MI_increase}.
\end{proof}

\begin{lemma}[Task MI and conditional entropy]
\label{lem:MI_entropy}
If $I(Y;\hat{\mathcal{T}})\ge I(Y;\mathcal{T})$, then
\begin{equation}
    H(Y\mid \hat{\mathcal{T}}) \;\le\; H(Y\mid \mathcal{T}).
    \label{eq:entropy_decrease}
\end{equation}
\end{lemma}

\begin{proof}
By the identity $I(Y;U) = H(Y) - H(Y\mid U)$ for any random variable $U$,
\[
H(Y\mid U) = H(Y) - I(Y;U).
\]
Thus
\[
H(Y\mid \hat{\mathcal{T}}) = H(Y) - I(Y;\hat{\mathcal{T}}),
\quad
H(Y\mid \mathcal{T}) = H(Y) - I(Y;\mathcal{T}).
\]
If $I(Y;\hat{\mathcal{T}})\ge I(Y;\mathcal{T})$, then
$H(Y\mid \hat{\mathcal{T}})\le H(Y\mid \mathcal{T})$, which is~\eqref{eq:entropy_decrease}.
\end{proof}

\begin{lemma}[Conditional entropy and confidence]
\label{lem:entropy_confidence}
Under Assumption~B.3, if 
$H(Y\mid \hat{\mathcal{T}})\le H(Y\mid \mathcal{T})$, then
\begin{equation}
    J_{\mathrm{conf}}(\hat{\mathcal{T}})
    \;\ge\;
    J_{\mathrm{conf}}(\mathcal{T}).
    \label{eq:conf_increase}
\end{equation}
\end{lemma}

\begin{proof}
By~\eqref{eq:B3_psi},
\[
J_{\mathrm{conf}}(\hat{\mathcal{T}})
= \psi\!\big(H(Y\mid \hat{\mathcal{T}})\big),
\quad
J_{\mathrm{conf}}(\mathcal{T})
= \psi\!\big(H(Y\mid \mathcal{T})\big).
\]
Since $\psi$ is strictly decreasing and
$H(Y\mid \hat{\mathcal{T}})\le H(Y\mid \mathcal{T})$,
we obtain~\eqref{eq:conf_increase}.
\end{proof}

\begin{theorem}[For Theorem 4.2]
\label{thm:theorem42_formal}
Under Assumptions~B.1--B.3, if
\begin{equation}
    I(\hat{\mathcal{T}}; z_v)\;\ge\; I(\mathcal{T}; z_v),
\end{equation}
then
\begin{equation}
    J_{\mathrm{conf}}(\hat{\mathcal{T}})
    \;\ge\;
    J_{\mathrm{conf}}(\mathcal{T}).
\end{equation}
\end{theorem}

\begin{proof}
From Assumption~B.1 and Lemma~\ref{lem:MI_to_task},
$I(\hat{\mathcal{T}}; z_v)\ge I(\mathcal{T}; z_v)$
implies $I(Y;\hat{\mathcal{T}})\ge I(Y;\mathcal{T})$.
By Lemma~\ref{lem:MI_entropy}, this yields
$H(Y\mid \hat{\mathcal{T}})\le H(Y\mid \mathcal{T})$.
Finally, Lemma~\ref{lem:entropy_confidence} implies
$J_{\mathrm{conf}}(\hat{\mathcal{T}})\ge J_{\mathrm{conf}}(\mathcal{T})$.
\end{proof}

\noindent Theorem~\ref{thm:theorem42_formal} formalizes that, under mild assumptions,
DVI-induced increases in the mutual information between latent states and visual
features translate into lower predictive uncertainty about the target output,
and thus into higher expected confidence.

\subsection{Derivation of the Latent Policy Gradient and Update Rule}
\label{app:latent_pg_full}
We give a detailed derivation of the gradient used to update the latent
thought vectors $H$ (e.g., latent think tokens) via a Gaussian policy gradient method.
\vspace{-0.2cm}
\paragraph{Objective.}
Let $A$ denote the sampled latent action (a perturbed latent thought
vector) and $R(A)$ the corresponding reward or confidence score. Given a
stochastic policy $\pi(A\mid H)$ over actions $A$ conditioned on the
current latent vector $H$, we define the expected objective
\begin{equation}
    J(H)
    := \mathbb{E}_{A\sim\pi(\cdot\mid H)}[R(A)].
    \label{eq:latent_obj}
\end{equation}

Our goal is to derive $\nabla_H J(H)$ and obtain a practical estimator
for use in gradient ascent.

\subsubsection{Policy Gradient Form}
\begin{lemma}[Policy gradient identity]
\label{lem:pg_identity_H}
Under mild regularity assumptions (allowing differentiation under the
integral sign), the gradient of $J(H)$ in~\eqref{eq:latent_obj} satisfies
\begin{equation}
    \nabla_H J(H)
    = \mathbb{E}_{A\sim\pi(\cdot\mid H)}
      \big[ R(A)\,\nabla_H \log \pi(A\mid H) \big].
    \label{eq:pg_theorem_H}
\end{equation}
\end{lemma}

\begin{proof}
By definition of expectation,
\begin{align}
    J(H)
    &= \int R(A)\,\pi(A\mid H)\,dA.
\end{align}
Differentiating w.r.t.\ $H$ yields
\begin{align}
    \nabla_H J(H)
    &= \int R(A)\,\nabla_H \pi(A\mid H)\,dA.
\end{align}
Using the log-derivative identity
$\nabla_H \pi(A\mid H) = \pi(A\mid H)\,\nabla_H \log\pi(A\mid H)$, we obtain
\begin{align}
    \nabla_H J(H)
    &= \int R(A)\,\pi(A\mid H)\,\nabla_H \log\pi(A\mid H)\,dA \\
    &= \mathbb{E}_{A\sim\pi(\cdot\mid H)}
       \big[ R(A)\,\nabla_H \log \pi(A\mid H) \big],
\end{align}
which is~\eqref{eq:pg_theorem_H}.
\end{proof}

\subsubsection{Gaussian Latent Policy}
We now instantiate the policy as an isotropic Gaussian perturbation of
$H$. Specifically, we sample
\begin{equation}
    A = H + \xi^{(t)},
    \qquad
    \xi^{(t)} \sim \mathcal{N}(0,\sigma^{2}I),
    \label{eq:gaussian_policy_A}
\end{equation}
which is equivalent to
\begin{equation}
    \pi(A\mid H)
    = \mathcal{N}(A;\,H,\sigma^{2}I).
\end{equation}

\noindent We next compute the gradient of the log-probability under this Gaussian
policy.

\begin{lemma}[Gradient of Gaussian log-probability]
\label{lem:gaussian_score}
For the Gaussian policy in~\eqref{eq:gaussian_policy_A}, the
log-probability gradient satisfies
\begin{equation}
    \nabla_H \log \pi(A\mid H)
    = \frac{A - H}{\sigma^{2}}
    = \frac{\xi}{\sigma^{2}}.
    \label{eq:gaussian_score}
\end{equation}
\end{lemma}

\begin{proof}
The log-density of $\pi(A\mid H)$ is
\begin{equation}
    \log\pi(A\mid H)
    = -\frac{1}{2\sigma^{2}}\|A-H\|^{2} + \mathrm{const},
\end{equation}
where the constant does not depend on $H$. Differentiating w.r.t.\ $H$,
\begin{align}
    \nabla_H \log\pi(A\mid H)
    &= -\frac{1}{2\sigma^{2}}
       \nabla_H \|A-H\|^{2} \\
    &= -\frac{1}{2\sigma^{2}}
       \nabla_H (A-H)^\top(A-H) \\
    &= -\frac{1}{2\sigma^{2}}\cdot(-2)(A-H) \\
    &= \frac{A-H}{\sigma^{2}}.
\end{align}
Using $A = H + \xi$ from~\eqref{eq:gaussian_policy_A} yields
$A-H = \xi$, giving~\eqref{eq:gaussian_score}.
\end{proof}

\noindent  Substituting~\eqref{eq:gaussian_score} into the policy gradient
expression~\eqref{eq:pg_theorem_H}, we obtain
\begin{align}
    \nabla_H J(H)
    &= \mathbb{E}_{A\sim\pi(\cdot\mid H)}
       \big[ R(A)\,\nabla_H \log\pi(A\mid H) \big] \\
    &= \mathbb{E}_{\epsilon\sim\mathcal{N}(0,\sigma^{2}I)}
       \left[
           R(H+\xi)\,\frac{\xi}{\sigma^{2}}
       \right].
    \label{eq:latent_pg_gaussian}
\end{align}

\subsubsection{Monte Carlo Estimator from a Single Sample}

In practice, the expectation in~\eqref{eq:latent_pg_gaussian} is
approximated via Monte Carlo sampling. At optimization step $t$, we draw
a single Gaussian noise sample
\[
\mathcal{\xi}^{(t)}\sim \mathcal{N}(0,\sigma^{2}I),
\qquad
A^{(t)} = H^{(t)} + \xi^{(t)},
\]
and construct the one-sample estimator
\begin{equation}
    \nabla_H J(H^{(t)})
    \approx R\big(H^{(t)} + \xi^{(t)}\big)
             \frac{\xi^{(t)}}{\sigma^{2}}.
    \label{eq:latent_pg_MC}
\end{equation}

\begin{lemma}[Unbiasedness of the Monte Carlo estimator]
\label{lem:unbiased_MC}
Let $\mathcal{\xi}^{(t)}\sim \mathcal{N}(0,\sigma^{2}I)$ and define the
estimator
\(
    \widehat{g}^{(t)}
    = R(H^{(t)} + \xi^{(t)})
      \xi^{(t)}/\sigma^{2}.
\)
Then
\begin{equation}
    \mathbb{E}[\widehat{g}^{(t)}]
    = \nabla_H J(H^{(t)}),
\end{equation}
where the expectation is taken over $\xi^{(t)}$.
\end{lemma}

\begin{proof}
By definition and linearity of expectation,
\begin{align}
    \mathbb{E}[\widehat{g}^{(t)}]
    &= \mathbb{E}_{\xi^{(t)}}
       \left[
           R(H^{(t)} + \xi^{(t)})
           \frac{\epsilon^{(t)}}{\sigma^{2}}
       \right]
     = \nabla_H J(H^{(t)}),
\end{align}
where the last equality is exactly
\eqref{eq:latent_pg_gaussian} evaluated at $H = H^{(t)}$.
\end{proof}

\subsubsection{Gradient Ascent Update}

Finally, we perform standard gradient ascent on the latent vector $H$.
With learning rate $\eta>0$, the update at step $t$ is
\begin{equation}
    H^{(t+1)}
    = H^{(t)}
      + \eta \cdot
        R\big(H^{(t)} + \xi^{(t)}\big)
        \frac{\xi^{(t)}}{\sigma^{2}}.
    \label{eq:latent_update_rule}
\end{equation}

\noindent This update moves the latent thought vector $H$ in the direction of
noisy perturbations $\xi^{(t)}$ that lead to higher reward, as
estimated by the Monte Carlo policy gradient.

\section{More detail about Visual Injection.}
\label{e}

\textbf{Different Strategies Discussion.} We explore two visual injection mechanisms to identify which strategy best supports multi-step latent reasoning optimization. \textbf{\textit{The first strategy}}, ultimately adopted in our DVI Strategy is a dynamic visual patch search-and-refinement procedure organized around the three-phase cycle shown in Algorithm~\ref{latent_reasoning}. At the start of the reasoning trajectory, we use the initial latent think token’s cross-modal attention to select the top-$m$ visually relevant patches, forming an initial candidate set $\mathcal{Z}_{\text{cand}}$ and an initial best-patch set $\mathcal{V}_{\text{best}}$. This initialization provides a reasonable but unconverged visual hypothesis, ensuring that the search begins from semantically meaningful regions while preserving room for subsequent refinement.

During each latent-update iteration, we recompute attention based on the updated hidden state $\mathcal{T}^{(t)}$ and resample a fresh batch of candidate patches $\mathcal{Z}_{\text{cand}} = \{Z_1, \dots, Z_m\}$. This dynamic resampling allows the visual hypothesis to adapt in tandem with the evolving reasoning context, preventing the model from becoming locked to the potentially imperfect initial attention pattern. We then construct an expanded latent representation $\tilde{\mathcal{T}}^{(t)}_l = [\mathcal{T}^{(t)}_l, \mathcal{Z}_{\text{cand}}, \mathcal{V}_{\text{best}}]$ and compute its reward $r = \mathcal{R}(\mathcal{Z}, \mathcal{Q}, \tilde{\mathcal{T}}^{(t)}_l)$, which acts as a task-level criterion for validating whether the newly sampled patches truly enhance reasoning performance.

If the obtained reward surpasses the historical best value $r_{\text{best}}$, we incorporate the new patches into the best-patch set $\mathcal{V}_{\text{best}}$ and carry forward the updated latent representation into the next iteration; otherwise, we revert the update to preserve previously verified visual evidence. Over multiple iterations, $\mathcal{V}_{\text{best}}$ is progressively refined under the interplay of exploration (resampling new patches) and exploitation (reward-based filtering), ultimately converging to a compact and semantically stable set of visual anchors. This dynamic best-patch optimization provides consistent and reward-validated visual grounding throughout the reasoning chain, producing far more reliable behavior than strategies that rely solely on per-step attention signals.

\textbf{\textit{The second strategy}} adopts a more direct formulation: at each iteration, it simply injects the visual patch that receives the highest attention, thereby emulating the attention-guided visual grounding commonly used in existing multimodal models. These two strategies thus represent fundamentally different paradigms for visual injection—one progressively optimizes visual evidence through reward-driven refinement, whereas the other relies on instantaneous attention-based selection. \textit{Given their distinct underlying principles, it is necessary to systematically compare their behaviors and performance characteristics.}

\noindent \textbf{Results Analysis.} As shwon in Table~\ref{tab:stability_compare} and \ref{tab:perf_compare}, we observe that injecting the highest attention visual patch at each iteration appears straightforward but suffers from fundamental instability. Attention scores are highly local and easily influenced by visual saliency, token co-adaptation, and early stage noise, causing the most attended region to deviate from the truly task relevant area. Because this strategy depends entirely on instantaneous attention distributions, the injected visual evidence fluctuates across iterations, resulting in drifting visual focus and compounding errors when early attention is misaligned. In contrast, our dynamic visual injection strategy, based on iterative patch resampling, reward evaluation, and best patch refinement, provides a far more stable grounding mechanism. By validating new patches through task level reward rather than noisy attention signals, the model progressively converges to a semantically consistent and task relevant set of visual anchors. This search, verify, refine process stabilizes visual grounding across iterations, prevents error amplification, and yields markedly more reliable reasoning trajectories. Empirically, the dynamic best patch optimization consistently outperforms attention based injection, demonstrating superior robustness and significantly higher multimodal reasoning accuracy.

\begin{table}[h]
\centering
\caption{Performance comparison between the two visual injection strategies on the MMVP Benchmark.}
\vspace{0.1cm}
\resizebox{0.8\linewidth}{!}{
\begin{tabular}{l c}
\toprule
\textbf{Method} & \textbf{Accuracy $\uparrow$} \\
\midrule
Attention-based  & 66.3 \\
Dynamic best-patch (ours) & \textbf{74.8} \\
\bottomrule
\end{tabular}}
\label{tab:perf_compare}
\end{table}

\begin{table}[h]
\centering
\caption{Stability metrics across iterative reasoning on the MathVision Benchmark. Lower is better.}
\vspace{0.1cm}
\resizebox{\linewidth}{!}{
\begin{tabular}{l cc}
\toprule
\textbf{Method} & \textbf{Attention Variance $\downarrow$} & \textbf{Drift Distance $\downarrow$} \\
\midrule
Attention-based  & 0.48 & 12.3 \\
Dynamic best-patch (ours) & \textbf{0.12} & \textbf{2.1} \\
\bottomrule
\end{tabular}}
\label{tab:stability_compare}
\end{table}

\noindent \textbf{In-depth Analysis.} When designing DVI, a seemingly reasonable question arises: since visual injection is intended to supply additional evidence when the model becomes uncertain, can confidence be used as a trigger signal so that visual patch optimization is performed only when confidence falls below a predefined threshold? This confidence-triggered variant appears more economical and aligns with the intuition of “inject vision only when needed.” However, our experiments (as shown in Table~\ref{tab:confidence_vs_continuous}) reveal that this strategy leads to noticeably worse convergence behavior. The optimization of visual patches in DVI is inherently continuous and incremental, relying on multi-step accumulation rather than sporadic corrections. Restricting updates to low confidence moments breaks this continuity. In practice, the model often enters a suboptimal regime in which confidence remains superficially stable while the underlying visual grounding is still misaligned, delaying or even preventing necessary corrections. As a result, the overall number of iterations required for convergence increases, and the final visual anchor becomes less stable and less semantically consistent than under the continuous best-patch refinement strategy. These findings indicate that confidence is not a reliable indicator for when visual refinement should occur, and maintaining per-iteration optimization is essential for achieving robust, efficient, and semantically coherent convergence in DVI.

\begin{table}[t]
\centering
\caption{
Comparison between confidence-triggered refinement and continuous DVI optimization on the MMVP Benchmark.
\textbf{Iter.} refers to the number of iterations required for convergence,
\textbf{Cons.} refers to the cross-iteration patch consistency,
\textbf{Err.} refers to the grounding error,
and \textbf{Acc.} refers to the final answer accuracy.
Continuous optimization converges faster, produces more stable visual anchors, and achieves higher accuracy.
}
\vspace{0.15cm}
\resizebox{\linewidth}{!}{
\begin{tabular}{lcccc}
\toprule
\textbf{Method} &
\textbf{Iter. $\downarrow$} &
\textbf{Cons. $\uparrow$} &
\textbf{Err. $\downarrow$} &
\textbf{Acc. $\uparrow$} \\
\midrule
Confidence-triggered update & 14.2 & 0.46 & 0.27 & 68.1 \\
Continuous best-patch (ours) & \textbf{9.3} & \textbf{0.83} & \textbf{0.12} & \textbf{74.8} \\
\bottomrule
\end{tabular}}
\label{tab:confidence_vs_continuous}
\vspace{-0.2cm}
\end{table}

\begin{algorithm}[h] 
\caption{\small \textbf{\textit{Dynamic Multimodal Latent Reasoning}}}
\small
\label{latent_reasoning}
\DontPrintSemicolon
\SetAlgoVlined
\SetKwProg{Fn}{}{}{}
\KwRequire{Image embeddings $\mathcal{Z}$, text embeddings $\mathcal{Q}$, 
latent tokens $\mathcal{T}_l$, learning rate $\eta$, 
iterations $T$, best visual patch $\mathcal{V}_{\textit{best}}$, 
top-$k$ probability $\mathrm{Top}_k(\mathcal{P}_i)$, the number of candidate patches 
$m$}

\vspace{2pt}
$\mathrm{Top}_k(\mathcal{P}_i) = \pi_{\theta}([\mathcal{Q}, \mathcal{Z}, \mathcal{T}])$; 
$r \gets \mathcal{R}(\mathcal{P}_{i})$ 
\hspace{1.em}\textcolor{blue}{$\triangleright$ reward}

\vspace{3pt}
\textcolor{darkgray}{\# \textit{Latent Policy Gradient Optimization}}\;

\For{$T = 1 \dots t$}{
    $\epsilon \sim \mathcal{N}(0,\sigma^2 I)$ 
    \hfill \textcolor{blue}{$\triangleright$ latent perturbation}

    $\mathcal{T}^{(t)\prime} \gets \mathcal{T}^{(t)} + \epsilon$

    $\mathcal{T}^{(t)} \gets \mathcal{T}^{(t)} + 
    \eta \nabla_{\mathcal{T}^{(t)}} \mathcal{J}(\mathcal{T}^{(t)})$
    \hfill \textcolor{blue}{$\triangleright$ latent update}

    \vspace{2pt}
    \textcolor{darkgray}{\# \textit{Dynamic Visual Injection}}\;
    
    $\mathcal{V}_{\textit{best}} \gets \text{Initialize}(\mathcal{T}^{(0)}, m)$ \hfill \textcolor{blue}{$\triangleright$ initialize best patch}

    \For{$L = 1 \dots l$}{
    
       $\mathcal{Z}_{\textit{cand}} \gets 
        \text{AttentionSelect}(\mathcal{T}_l^{(t)}, m)$
        \hfill \textcolor{blue}{$\triangleright$ select $m$ candidate visual patches}

        $\tilde{\mathcal{T}_l}^{(t)} \gets [\mathcal{T}_l^{(t)}, \mathcal{Z}_{\textit{cand}}, \mathcal{V}_{\textit{best}}]$

        $r \gets \mathcal{R}(\mathcal{Q}, \mathcal{Z}, \tilde{\mathcal{T}^{(t)}_l})$

        \If{$r > r_{\text{best}}$}{
            $\mathcal{V}_{\textit{best}} \gets \mathcal{V}_{\textit{best}} \cup \mathcal{Z}_{\textit{cand}}$;

            $\mathcal{T}_l^{(t)} \gets \tilde{\mathcal{T}_l}^{(t)}$
            \hfill \textcolor{blue}{$\triangleright$ update best}
        }
        \Else{
            $\mathcal{T}_l^{(t)} \gets 
            [\mathcal{T}_l^{(t)}, \mathcal{V}_{\textit{best}}]$
            \hfill \textcolor{blue}{$\triangleright$ revert to previous best}
        }
        \vspace{1pt}
    }
    \vspace{2pt}
}

\vspace{3pt}
$\mathcal{X} \gets \text{Decode}(\mathcal{T}^{(t)}, \mathcal{Z}, \mathcal{Q})$

\Return $\mathcal{X}$
\end{algorithm}

\section{Case Study}
\label{f}
In this section, we provide qualitative examples to illustrate the behavior of the DMLR-enhanced model across different task settings. As shown in \textbf{Example 1-8}, the model demonstrates more reliable and fine-grained visual perception, supported by stable and semantically coherent internal representations. In reasoning scenarios, the model produces accurate and well-structured logical steps. Importantly, the dynamic latent reasoning mechanism suppresses unnecessary expansions, resulting in concise and task-focused reasoning chains rather than overly long or speculative explanations. Taken together, these examples highlight that DMLR simultaneously strengthens perceptual quality and reasoning reliability, while ensuring that the generated reasoning remains efficient and free from redundant elaboration.

\begin{figure*}[t]
    \centering
    \begin{mdframed}[
        frametitle={Perception Example 1},
        frametitlebackgroundcolor=lightyellow,
        backgroundcolor=gray!10,
        linecolor=gray,
        frametitlerule=true,
        frametitlefont=\color{black}\bfseries
    ]
        \centering
        \small
        \includegraphics[width=0.4\linewidth]{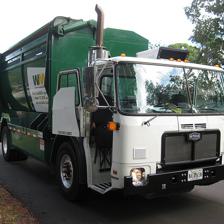} 
        \captionof{figure}{\centering Question: Is the door of the truck cab open?\\Options: (a) Yes (b) No\\Ground Truth: (a)       \label{fig1}} 
  
        \flushleft

    \noindent\textbf{Qwen2.5-VL-7B-Instruct + DMLR:}

    The image shows a garbage truck with its door on the side of the cab open. This is evident from the visible interior and the fact that the door is not aligned with the rest of the vehicle's structure, indicating it has been opened. The answer is \boxed{a}. 

    \vspace{1em}

    \noindent\textbf{Qwen2.5-VL-7B-Instruct:}

    To determine if the door of the truck cab is open, let's analyze the image step by step:
    \begin{enumerate}
        \item Observe the side of the truck where the door would be located.
        \item Look for any visible gap or opening that would indicate the door is not closed.
        \item Check the overall appearance of the door; if it is open, there should be a noticeable gap or difference in the alignment compared to when it is closed.
    \end{enumerate}

    Upon examining the image:
    \begin{itemize}
        \item The door appears to be aligned with the rest of the truck's body.
        \item There is no visible gap or difference in the alignment that would suggest the door is open.
    \end{itemize}

    Based on this analysis, the door of the truck cab does not appear to be open.

    \noindent\texttt{<think>} The door of the truck cab is not open.\texttt{</think>}
    \noindent\texttt{<answer>} No\texttt{</answer>}

    \noindent\boxed{\text{No}}
       \label{fig1}
    \end{mdframed}
\end{figure*}

\begin{figure*}[t]
    \centering
    \begin{mdframed}[
        frametitle={Perception Example 2},
        frametitlebackgroundcolor=lightyellow,
        backgroundcolor=gray!10,
        linecolor=gray,
        frametitlerule=true,
        frametitlefont=\color{black}\bfseries
    ]
        \centering
        \small
        \includegraphics[width=0.4\linewidth]{img/example/mmvp2.jpg} 
        \captionof{figure}{\centering Question: What is the woman more likely to do?\\Options: (a) Bump (b) Set\\Ground Truth: (b)} 
        \flushleft

    \noindent\textbf{Qwen2.5-VL-7B-Instruct + DMLR:}

    In the image, a volleyball game is in progress. The woman in focus appears to be positioned near the net, likely preparing to receive the ball. In volleyball, there are two primary actions for receiving the ball: bumping and setting.

    \begin{itemize}
        \item Bumping involves using the hands to cushion the ball, typically when it's coming down from above.
        \item Setting involves using the hands to lift the ball upward, usually when it's coming down or is at a low trajectory.
    \end{itemize}

    Given her position and the context of the game, she seems to be preparing to set the ball rather than bump it. This is because setting is a more strategic action that allows the team to control the direction and height of the ball, which is crucial for passing and spiking effectively.

    Therefore, based on the context and typical volleyball gameplay, the woman is more likely to set the ball. The answer is \boxed{b) \text{Set}}.

    \vspace{1em}

    \noindent\textbf{Qwen2.5-VL-7B-Instruct:}

    To determine what the woman is more likely to do, let's analyze the image step by step:
    \begin{enumerate}
        \item \textbf{Identify the context}: The image shows a volleyball game in progress. The woman in the foreground appears to be preparing to receive the ball, which is above her head.
        \item \textbf{Understand the volleyball terms}:
        \begin{itemize}
            \item \textbf{Bumping}: This is when a player uses their hands to pass the ball back into play after it has been spiked or set.
            \item \textbf{Setting}: This is when a player uses their hands to pass the ball high into the air for a teammate to spike.
        \end{itemize}
        \item \textbf{Analyze the woman's position}:
        \begin{itemize}
            \item The ball is above her head, and she appears to be in a position where she is about to use her hands to pass the ball.
            \item Her body posture suggests she is preparing to use her hands to control the ball, which is characteristic of bumping rather than setting.
        \end{itemize}
        \item \textbf{Consider the options}:
        \begin{itemize}
            \item Option (a) Bump: This aligns with her posture and the context of the game.
            \item Option (b) Set: This would require her to position herself differently and use her hands in a way that is not consistent with her current stance.
        \end{itemize}
    \end{enumerate}

    Given the analysis, the woman is more likely to bump the ball.

    \noindent\texttt{<think>} The woman is in a position to use her hands to control the ball, which is characteristic of bumping rather than setting.\texttt{</think>}
    \noindent\texttt{<answer>} \boxed{a} \texttt{</answer>}

    \end{mdframed}
\end{figure*}

\begin{figure*}[t]
    \centering
    \begin{mdframed}[
        frametitle={Perception Example 3},
        frametitlebackgroundcolor=lightyellow,
        backgroundcolor=gray!10,
        linecolor=gray,
        frametitlerule=true,
        frametitlefont=\color{black}\bfseries
    ]
        \centering
        \small
        \includegraphics[width=0.4\linewidth]{img/example/mmvp3.jpg} 
        \captionof{figure}{\centering Question: Can you see the caudal fin of the shark in the image?\\Options: (a) Yes (b) No\\Ground Truth: (a) Yes} 
        \flushleft

    \noindent\textbf{VLAA-Thinker-Qwen2.5VL-7B + DMLR:}
    To determine if the caudal fin of the shark is visible in the image, I need to analyze the orientation and positioning of the shark. The caudal fin is typically located at the rear end of the shark's body. In this image, the shark is shown from a side angle, and its tail is clearly visible extending towards the bottom right corner of the image. This indicates that the caudal fin can be seen.

    Therefore, based on the visibility of the tail in the image, the answer is (a) Yes. The answer is \boxed{a}.

    \vspace{1em}

    \noindent\textbf{VLAA-Thinker-Qwen2.5VL-7B:}

    \texttt{<think>}
    To determine if the caudal fin of the shark is visible in the image, I need to analyze the orientation and position of the shark. The caudal fin is typically located at the rear end of the shark's body. In this image, the shark is shown from a side angle, and the caudal fin is not directly visible due to the angle and the positioning of the shark's body. However, the tail area, which includes the caudal fin, can be inferred to be present based on the overall shape and structure of the shark.

    Given the options:
    (a) Yes
    (b) No

    The caudal fin is not directly visible in the image, but it can be inferred to be present based on the shark's anatomy and the angle of the image.

    Therefore, the correct answer is (b) No.
    \texttt{</think>}

    \noindent\texttt{<answer>} (b) No \texttt{</answer>}

    \end{mdframed}
\end{figure*}

\begin{figure*}[t]
    \centering
    \begin{mdframed}[
        frametitle={Perception Example 4},
        frametitlebackgroundcolor=lightyellow,
        backgroundcolor=gray!10,
        linecolor=gray,
        frametitlerule=true,
        frametitlefont=\color{black}\bfseries
    ]
        \centering
        \small
        \includegraphics[width=0.4\linewidth]{img/example/mmvp4.jpg} 
        \captionof{figure}{\centering Question: Is the flame of the match more round or thin?\\Options: (a) Round (b) Thin\\Ground Truth: (b) Thin} 
        \flushleft

    \noindent\textbf{VLAA-Thinker-Qwen2.5VL-7B + DMLR:}

    To determine whether the flame of the match is more round or thin, let's analyze the image. The flame appears elongated and narrow at the top, which suggests it is not round but rather thin in width. The base of the flame seems wider, but the overall shape of the flame is more elongated than circular.

    Therefore, based on the visual analysis, the flame of the match is more thin. The answer is \boxed{b}.

    \vspace{1em}

    \noindent\textbf{VLAA-Thinker-Qwen2.5VL-7B:}

    \texttt{<think>}
To determine whether the flame of the match is more round or thin, let's analyze the image. The flame appears to have a distinct shape where the base is wider and narrows towards the top, which is characteristic of a conical shape. This shape suggests that the flame is not uniformly round but rather has a tapering effect from bottom to top. Given this observation, the flame can be described as more round at its base and becoming thinner towards the top, but overall, it maintains a rounded appearance.

Therefore, based on the visual analysis, the flame of the match is more round.
    \texttt{</think>}

    \noindent\texttt{<answer>} \boxed{a} \texttt{</answer>}

    \end{mdframed}
\end{figure*}

\begin{figure*}[t]
    \centering
    \begin{mdframed}[
        frametitle={Reasoning Example 1},
        frametitlebackgroundcolor=lightyellow,
        backgroundcolor=gray!10,
        linecolor=gray,
        frametitlerule=true,
        frametitlefont=\color{black}\bfseries
    ]
        \centering
        \small
        \includegraphics[width=0.4\linewidth]{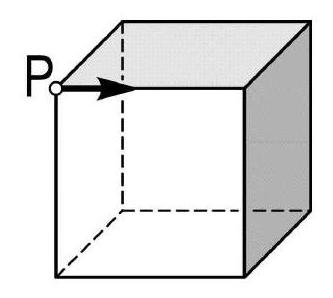} 
        \captionof{figure}{\centering Question: A beetle walks along the edges of a cube. Starting from point $P$ it first moves in the direction shown. At the end of each edge it changes the direction in which it turns, turning first right then left, then right etc. Along how many edges will it walk before it returns to point $P$?\\Ground Truth: 6} 
        \flushleft

    \noindent\textbf{Qwen2.5-VL-7B-Instruct + DMLR:}

    The beetle starts at point P and moves along the edges of a cube, changing direction according to the pattern: right, left, right. To determine how many edges it will walk before returning to point P, we need to trace its path.

    \begin{enumerate}
        \item The beetle starts at P.
        \item It moves right to an adjacent vertex.
        \item It turns left to another adjacent vertex.
        \item It turns right again to reach a third adjacent vertex.
        \item It continues this pattern until it returns to P.
    \end{enumerate}

    By examining the cube's structure, we can see that after 6 edges, the beetle will return to point P because each face of the cube has 4 edges, and the beetle covers all edges of one face before repeating the pattern on the next face.

    Thus, the beetle will walk along 6 edges before returning to point P.

    \vspace{1em}

    \noindent\textbf{Qwen2.5-VL-7B-Instruct:}

    To solve this problem, we need to track the beetle's path as it moves along the edges of the cube, changing its direction according to the specified pattern (right, left, right, etc.).

    \begin{enumerate}
        \item \textbf{Initial Position and Direction}: The beetle starts at point $P$ and moves in the direction indicated by the arrow, which is along one edge of the cube.
        \item \textbf{First Edge}: The beetle moves along the first edge and changes direction to the right.
        \item \textbf{Second Edge}: Moving right, the beetle now faces a new direction and continues along the next edge.
        \item \textbf{Third Edge}: After moving along the second edge, the beetle changes direction again to the left.
        \item \textbf{Fourth Edge}: Moving left, the beetle now faces another new direction and continues along the next edge.
        \item \textbf{Fifth Edge}: After moving along the third edge, the beetle changes direction to the right again.
        \item \textbf{Sixth Edge}: Moving right, the beetle now faces yet another new direction and continues along the next edge.
        \item \textbf{Seventh Edge}: After moving along the fourth edge, the beetle changes direction to the left once more.
        \item \textbf{Eighth Edge}: Moving left, the beetle now faces a new direction and continues along the next edge.
    \end{enumerate}

    At this point, the beetle has completed a full cycle around the cube, returning to point $P$.

    The beetle will have walked along 8 edges before it returns to point $P$. The answer is 8.

    \end{mdframed}
\end{figure*}

\begin{figure*}[p]
    \centering
    \begin{mdframed}[
        frametitle={Reasoning Example 2},
        frametitlebackgroundcolor=lightyellow,
        backgroundcolor=gray!10,
        linecolor=gray,
        frametitlerule=true,
        frametitlefont=\color{black}\bfseries
    ]
        \centering
        \small
        \includegraphics[width=0.4\linewidth]{img/example/mathvision2.png} 
        \captionof{figure}{\centering Question: In the adjoining plane figure, sides $AF$ and $CD$ are parallel, as are sides $AB$ and $EF$, and sides $BC$ and $ED$. Each side has length of 1. Also, $\measuredangle FAB = \measuredangle BCD = 60^\circ$. The area of the figure is\\Options: A. $\frac{\sqrt{3}}{2}$, B. $1$, C. $\frac{3}{2}$, D. $\sqrt{3}$, E. $2$\\Ground Truth: D. $\sqrt{3}$} 
        \flushleft

    \noindent\textbf{Qwen2.5-VL-7B-Instruct + DMLR:}

    The given figure is a hexagon with sides of equal length and specific angles. To find the area, we can break it down into simpler shapes like triangles and parallelograms.

    \begin{enumerate}
        \item \textbf{Identify the shape}: The hexagon can be divided into two congruent trapezoids by drawing a diagonal from one vertex to the opposite non-adjacent vertex.
        \item \textbf{Calculate the area of each trapezoid}:
        \begin{itemize}
            \item Each trapezoid has parallel sides of 1 unit (AF and CD), and the height can be calculated using the properties of the 60° angle.
            \item The height of the trapezoid can be found using the sine of 60° in a right triangle formed by dropping a perpendicular from one of the vertices to the base.
            \item The area of one trapezoid is $\frac{1}{2} \times (\text{sum of parallel sides}) \times \text{height}$.
        \end{itemize}
    \end{enumerate}

    The height $h$ of the trapezoid can be calculated as follows: $h = 1 \cdot \sin(60^\circ) = \frac{\sqrt{3}}{2}$.

    The area of one trapezoid is: $\text{Area of one trapezoid} = \frac{1}{2} \times (1 + 1) \times \frac{\sqrt{3}}{2} = \frac{1}{2} \times 2 \times \frac{\sqrt{3}}{2} = \frac{\sqrt{3}}{2}$.

    Since there are two such trapezoids, the total area of the hexagon is: $\text{Total Area} = 2 \times \frac{\sqrt{3}}{2} = \sqrt{3}$.

    \vspace{1em}

    \noindent\textbf{Qwen2.5-VL-7B-Instruct:}

    To find the area of the given figure, we need to recognize that the shape can be divided into two congruent trapezoids. Let's break down the problem step by step:

    \begin{enumerate}
        \item \textbf{Identify the Shape}: The given figure is a hexagon with specific angle and side lengths. Since $AF$ is parallel to $CD$, $AB$ is parallel to $EF$, and $BC$ is parallel to $ED$, the hexagon can be divided into two congruent trapezoids by drawing a diagonal from $A$ to $D$.
        \item \textbf{Calculate the Area of One Trapezoid}: Each trapezoid is formed by two right triangles and a rectangle in the middle. The key angles are $60^\circ$ at $FAB$ and $BCD$. This means that the trapezoid can be split into a rectangle and two 30-60-90 triangles.
        \item \textbf{Properties of 30-60-90 Triangles}: In a 30-60-90 triangle, the sides are in the ratio $1 : \sqrt{3} : 2$. Here, the shorter leg (opposite the $30^\circ$ angle) is 1, so the longer leg (opposite the $60^\circ$ angle) is $\sqrt{3}$, and the hypotenuse is 2.
        \item \textbf{Area Calculation}:
        \begin{itemize}
            \item The height of each trapezoid is the same as the shorter leg of the 30-60-90 triangle, which is 1.
            \item The bases of the trapezoid are the sum of the shorter legs of the two 30-60-90 triangles, which is $1 + 1 = 2$.
            \item The area of one trapezoid is given by the formula for the area of a trapezoid: $\text{Area} = \frac{1}{2} \times (\text{Base}_1 + \text{Base}_2) \times \text{Height}$.
            \item Substituting the values, we get: $\text{Area} = \frac{1}{2} \times (1 + 1) \times 1 = \frac{1}{2} \times 2 \times 1 = 1$.
        \end{itemize}
        \item \textbf{Total Area}: Since the hexagon is composed of two such congruent trapezoids, the total area is $1 + 1 = 2$.
    \end{enumerate}

    Therefore, the area of the figure is \boxed{2}.

    \end{mdframed}
\end{figure*} 

\begin{figure*}[p] 
    \centering
    \begin{mdframed}[
        frametitle={Reasoning Example 3 (Part 1)},
        frametitlebackgroundcolor=lightyellow,
        backgroundcolor=gray!10,
        linecolor=gray,
        frametitlerule=true,
        frametitlefont=\color{black}\bfseries
    ]
        \centering
        \small
        \includegraphics[width=0.4\linewidth]{img/example/mathvision3.png} 
        \captionof{figure}{\centering Question: The solid shown in the diagram has 12 regular pentagonal faces... (content omitted for brevity) ... Total of the numbers written on the solid?\\Ground Truth: 50} 
        \flushleft

    \noindent\textbf{VLAA-Thinker-Qwen2.5VL-7B + DMLR:}

    To solve this problem, we need to determine the number of triangular and square faces on the solid. Given that there are 12 pentagonal faces, each surrounded by 5 square faces, we can calculate the total number of square faces involved in these connections.

    Each pentagonal face is connected to 5 square faces, so for 12 pentagonal faces, the total count of square faces from pentagons alone would be $12 \times 5 = 60$. However, since each square face is shared between two pentagons, the actual number of unique square faces is half of this, which is $60 / 2 = 30$.

    Next, let's denote the number of triangular faces as $T$ and the number of square faces as $S$. We know: $S = 30$.
    The total number of faces is given by the sum of pentagonal, triangular, and square faces: $12 + T + 30 = F$, where $F$ is the total number of faces. Since each triangular face is surrounded by 3 square faces, and each square face is counted twice (once per triangle and once per pentagon), we can use the fact that each square face is part of 3 triangles or 2 pentagons to find $T$.

    Given the structure, we can derive: $3T = 2 \times 30 \implies 3T = 60 \implies T = 20$.

    Thus, the total number of faces $F$ is: $F = 12 + 20 + 30 = 62$.

    Now, calculating the total sum of numbers written on the solid:
    \begin{itemize}
        \item Each pentagonal face contributes $5$: $12 \times 5 = 60$
        \item Each triangular face contributes $1$: $20 \times 1 = 20$
        \item Each square face contributes $-1$: $30 \times -1 = -30$
    \end{itemize}

    Summing these contributions: $60 + 20 - 30 = 50$.

    Therefore, the total of the numbers written on the solid is $50$.

    \end{mdframed}
\end{figure*}

\begin{figure*}[t] 
    \centering
    \begin{mdframed}[
        frametitle={Reasoning Example 3 (Part 2)}, 
        frametitlebackgroundcolor=lightyellow,
        backgroundcolor=gray!10,
        linecolor=gray,
        frametitlerule=true,
        frametitlefont=\color{black}\bfseries
    ]
    \flushleft
    \small

    \noindent\textbf{VLAA-Thinker-Qwen2.5VL-7B:}

    \texttt{<think>}
    To solve this problem, we need to determine the number of each type of face (pentagon, triangle, and square) on the solid and then sum the values written on these faces according to the given rules.

    \begin{enumerate}
        \item \textbf{Identify the types of faces and their properties}:
        \begin{itemize}
            \item The solid has 12 pentagonal faces.
            \item Let $T$ be the number of triangular faces.
            \item Let $S$ be the number of square faces.
        \end{itemize}
        \item \textbf{Use the given information about the arrangement of faces}:
        \begin{itemize}
            \item Each pentagonal face is surrounded by 5 square faces.
            \item Each triangular face is surrounded by 3 square faces.
        \end{itemize}
        \item \textbf{Set up equations based on the surrounding faces}:
        \begin{itemize}
            \item Since each pentagonal face is surrounded by 5 square faces, and there are 12 pentagonal faces, the total contribution from pentagons to the count of square faces is $12 \times 5 = 60$.
            \item Since each triangular face is surrounded by 3 square faces, the total contribution from triangular faces to the count of square faces is $3T$.
        \end{itemize}
        \item \textbf{Since each square face is counted multiple times (once for each face it touches), we can set up the equation for the total number of square faces}:
        $60 + 3T = 4S$.
        This equation comes from the fact that each square face is part of one pentagonal face and $T$ triangular faces, contributing to the total count of square faces.
        \item \textbf{Determine the number of triangular faces $T$}: We know the total number of faces is $12 + T + S$. However, since each triangular face contributes 3 square faces and each pentagonal face contributes 5 square faces, we can use the relationship derived: $60 + 3T = 4S$.
        \item \textbf{Solve for $S$ in terms of $T$}: $4S = 60 + 3T \implies S = \frac{60 + 3T}{4}$.
        \item \textbf{Substitute back into the total face count equation}: $12 + T + S = 12 + T + \frac{60 + 3T}{4} = 12 + T + 15 + \frac{3T}{4} = 27 + \frac{7T}{4}$.
        \item \textbf{Solve for $T$ such that $S$ is an integer}:
        \begin{itemize}
            \item For $S$ to be an integer, $\frac{60 + 3T}{4}$ must be an integer. Testing values: $60 + 3T \equiv 0 \pmod{4} \implies 3T \equiv -60 \pmod{4} \implies 3T \equiv 0 \pmod{4} \implies T \equiv 0 \pmod{4}$.
            \item Trying $T = 4$: $S = \frac{60 + 3 \times 4}{4} = \frac{72}{4} = 18$.
        \end{itemize}
        \item \textbf{Verify the solution}:
        \begin{itemize}
            \item Total faces: $12 + 4 + 18 = 34$
            \item Square faces: $60 + 3 \times 4 = 72$ divided by 4: $18$
        \end{itemize}
        \item \textbf{Calculate the total value}:
        \begin{itemize}
            \item Pentagons: $12 \times 5 = 60$
            \item Triangles: $4 \times 1 = 4$
            \item Squares: $18 \times (-1) = -18$
            \item Total: $60 + 4 - 18 = 46$
        \end{itemize}
    \end{enumerate}

    Thus, the total of the numbers written on the solid is \boxed{46}.
    \texttt{</think>}

    \noindent\texttt{<answer>} \boxed{46} \texttt{</answer>}

    \end{mdframed}
\end{figure*}

\begin{figure*}[p]
    \centering
    \begin{mdframed}[
        frametitle={Reasoning Example 4 (Part 1)},
        frametitlebackgroundcolor=lightyellow,
        backgroundcolor=gray!10,
        linecolor=gray,
        frametitlerule=true,
        frametitlefont=\color{black}\bfseries
    ]
        \centering
        \small
        \includegraphics[width=0.4\linewidth]{img/example/mathvision4.png} 
        \captionof{figure}{\centering Question: We want to paint each square in the grid with the colours P, Q, R and S, so that neighbouring squares always have different colours. (Squares which share the same corner point also count as neighbouring.) Some of the squares are already painted. In which colour(s) could the grey square be painted?\\Options: A. only Q, B. only R, C. only S, D. either R or S, E. it is not possible.\\Ground Truth: D. either R or S} 
        \flushleft

    \noindent\textbf{VLAA-Thinker-Qwen2.5VL-7B + DMLR:}

    To determine the color of the grey square, we need to analyze the constraints given by the problem. The grid is filled with the colors P, Q, R, and S such that no two adjacent squares (including diagonally adjacent ones) share the same color.

    \begin{enumerate}
        \item \textbf{Identify Neighboring Squares}:
        \begin{itemize}
            \item The grey square has neighbors in the following positions:
            \item Top-left: Q
            \item Top-right: 
            \item Bottom-left: 
            \item Bottom-right: 
        \end{itemize}
        \item \textbf{Analyze the Top-Left Neighbor}:
        \begin{itemize}
            \item The top-left neighbor is already colored Q.
            \item Therefore, the grey square cannot be Q.
        \end{itemize}
        \item \textbf{Consider Possible Colors for the Grey Square}:
        \begin{itemize}
            \item Since the grey square cannot be Q, it must be either R or S.
            \item We need to check if both R and S are possible without violating the rule that no two adjacent squares can have the same color.
        \end{itemize}
        \item \textbf{Check for Validity of R}:
        \begin{itemize}
            \item If the grey square were R, then the top-right square would need to be a different color from R and Q. This leaves P as the only valid option for the top-right square since it must differ from both R and Q.
        \end{itemize}
        \item \textbf{Check for Validity of S}:
        \begin{itemize}
            \item If the grey square were S, then the bottom-right square would need to be a different color from S and Q. This leaves P as the only valid option for the bottom-right square since it must differ from both S and Q.
        \end{itemize}
        \item \textbf{Conclusion}:
        \begin{itemize}
            \item Both R and S are valid options for the grey square because they do not violate the adjacency rules when considering their neighboring squares.
        \end{itemize}
    \end{enumerate}

    Therefore, the grey square could be painted either R or S.

    \end{mdframed}
\end{figure*}

\begin{figure*}[t]
    \centering
    \begin{mdframed}[
        frametitle={Reasoning Example 4 (Part 2)},
        frametitlebackgroundcolor=lightyellow,
        backgroundcolor=gray!10,
        linecolor=gray,
        frametitlerule=true,
        frametitlefont=\color{black}\bfseries
    ]
        \flushleft
        \small

    \noindent\textbf{VLAA-Thinker-Qwen2.5VL-7B:}

    \texttt{<think>}
    To determine the color of the grey square, we need to consider the constraints given: no two neighboring squares can have the same color, including those that share a corner.

    \begin{enumerate}
        \item The square directly above the grey square is already colored 'Q'.
        \item The square to the left of the grey square is colored 'R'.
        \item The square below the grey square is colored 'Q'.
    \end{enumerate}

    Given these constraints:
    \begin{itemize}
        \item The grey square cannot be 'Q' because it shares a corner with the square above it which is 'Q'.
        \item The grey square cannot be 'R' because it shares a corner with the square to the left which is 'R'.
        \item The grey square cannot be 'P' because there is no information suggesting 'P' is used elsewhere in a way that would conflict with the rules for the grey square.
    \end{itemize}

    Therefore, the only remaining color for the grey square is 'S'.
    \texttt{</think>}

    \noindent\texttt{<answer>} \boxed{S} \texttt{</answer>}

    \end{mdframed}
    \label{fig8}
\end{figure*}

\section{Prompt}
\label{g}
In this section, we provide the complete set of prompts used in our experiments. These prompts cover all evaluation settings, including reasoning, perception, visual-grounding analysis, confidence-based studies, and ablation configurations. By releasing the full prompt formulations, we aim to ensure that every component of our experimental pipeline is transparent and fully reproducible, enabling researchers to faithfully replicate our results and conduct further analysis based on the same standardized inputs.

\label{prompt}
\begin{figure*}[t]
\centering
\begin{tcolorbox}[
    width=\textwidth,            
    colback=gray!10,
    colframe=gray,
    coltitle=white,
    fonttitle=\bfseries,
    title=Reasoning Chains Correct Judge,
    attach boxed title to top left,
    boxed title style={
        colback=gray!50!black,
        boxrule=0pt,
        arc=0pt,
        outer arc=0pt,
        top=3pt,
        bottom=3pt,
        left=4pt,
        right=4pt
    },
    enhanced,
    sharp corners,
    left=6pt,
    right=6pt,
    top=6pt,
    bottom=6pt,
    halign title=flush left,     
    before upper=\raggedright    
]
You are given a multimodal reasoning example.

-- The model receives an image and a question (possibly with answer options).\\
-- The model then produces a step-by-step reasoning chain, followed by a final answer.\\
-- You are also given the ground-truth correct answer.\\

Your task is to decide whether the model's final answer is correct or incorrect.

Please follow these rules:\\
1.~Ignore style and verbosity. Focus only on factual correctness.\\
2.~If the final answer matches the ground-truth answer, label it as \textbf{CORRECT}, even if some intermediate reasoning steps are slightly suboptimal.\\
3.~If the final answer does \textbf{NOT} match the ground-truth answer, label it as \textbf{INCORRECT}, even if parts of the reasoning look plausible.\\
4.~If the model outputs multiple conflicting answers, treat the final one as its final answer.\\

Return your decision in the following JSON format only:

\begin{verbatim}
{
  "label": "CORRECT" or "INCORRECT",
  "short_explanation": "<one-sentence>"
}
\end{verbatim}

\noindent QUESTION: \texttt{\{QUESTION\_TEXT\}}\\
OPTIONS (if any): \texttt{\{OPTION\_TEXT\}}\\
GROUND-TRUTH ANSWER: \texttt{\{GT\_ANSWER\}}\\
MODEL REASONING CHAIN: \texttt{\{MODEL\_CHAIN\}}\\
MODEL FINAL ANSWER: \texttt{\{MODEL\_FINAL\_ANSWER\}}

\end{tcolorbox}
\end{figure*}

\begin{figure*}[t]
\centering
\begin{tcolorbox}[
    width=\textwidth,
    colback=gray!10,
    colframe=gray,
    coltitle=white,
    fonttitle=\bfseries,
    title=Faithfulness Reasoning Judge,
    attach boxed title to top left,
    boxed title style={
        colback=gray!50!black,
        boxrule=0pt,
        arc=0pt,
        outer arc=0pt,
        top=3pt,
        bottom=3pt,
        left=4pt,
        right=4pt
    },
    enhanced,
    sharp corners,
    left=6pt,
    right=6pt,
    top=6pt,
    bottom=6pt,
    halign title=flush left,
    before upper=\raggedright
]
You are given a multimodal reasoning example.\par

The model receives:\par
- An image (you can refer to its description below), and\par
- A textual question (possibly with answer options).\par

The model then produces a step-by-step reasoning chain and a final answer.\par

Your task is to judge whether the reasoning chain is FAITHFUL to the information
in the question and image description, or UNFAITHFUL.\par

Definitions:\par
- A FAITHFUL chain explicitly uses relevant information from the question and image
  (or its description), and each step logically contributes to deriving the final answer.\par
- An UNFAITHFUL chain either:\par
  * ignores critical visual/textual details,\par
  * relies on generic templates or guesses,\par
  * contains logically inconsistent steps, or\par
  * gives a correct answer but with reasoning that does not match the actual evidence.\par

Instructions:\par
1. Check whether the reasoning steps are grounded in the given question and image description.\par
2. Check whether the logical steps are coherent and consistent with each other.\par
3. A chain can be UNFAITHFUL even if the final answer happens to be correct by coincidence.\par

Return your judgment in the following JSON format only:\par

\begin{verbatim}
{
  "label": "FAITHFUL" or "UNFAITHFUL",
  "short_explanation": "<one-sentence justification>"
}
\end{verbatim}

QUESTION:\par
\texttt{\{QUESTION\_TEXT\}}\par

IMAGE DESCRIPTION (if available):\par
\texttt{\{IMAGE\_DESCRIPTION\}}\par

MODEL REASONING CHAIN:\par
\texttt{\{MODEL\_CHAIN\}}\par

MODEL FINAL ANSWER:\par
\texttt{\{MODEL\_FINAL\_ANSWER\}}

\end{tcolorbox}
\end{figure*}

\begin{figure*}[t]
\centering
\begin{tcolorbox}[
    width=\textwidth,
    colback=gray!10,
    colframe=gray,
    coltitle=white,
    fonttitle=\bfseries,
    title=Visual Hallucination Judge,
    attach boxed title to top left,
    boxed title style={
        colback=gray!50!black,
        boxrule=0pt,
        arc=0pt,
        outer arc=0pt,
        top=3pt,
        bottom=3pt,
        left=4pt,
        right=4pt
    },
    enhanced,
    sharp corners,
    left=6pt,
    right=6pt,
    top=6pt,
    bottom=6pt,
    halign title=flush left,
    before upper=\raggedright
]
\small

You are given:\par
- An image,\par
- A question about the image, and\par
- Several reasoning statements produced by a model that refer to visual content.\par

Your task is to check whether each visual statement is supported by the image
(or its description), or whether it hallucinates objects, attributes, or relations
that do not actually appear in the image.\par

Definitions:\par
- A NON-HALLUCINATED (grounded) statement is consistent with the image:
  mentioned objects, attributes, and relations are all present or at least plausible.\par
- A HALLUCINATED statement mentions objects, attributes, numbers, or spatial relations
  that contradict the image or clearly do not appear in it.\par

Instructions:\par
1. Carefully compare each visual statement with the image (or its description).\par
2. For each statement, decide whether it is HALLUCINATED or NON-HALLUCINATED.\par
3. Ignore minor wording differences; focus on factual consistency.\par

Return your output in the following JSON format:\par

\begin{verbatim}
{
  "results": [
    {
      "statement_id": 1,
      "label": "HALLUCINATED" or "NON-HALLUCINATED",
      "short_explanation": "<short reason>"
    },
    {
      "statement_id": 2,
      "label": "...",
      "short_explanation": "..."
    }
    ...
  ]
}
\end{verbatim}

--------------------\par
QUESTION:\par
\texttt{\{QUESTION\_TEXT\}}\par

IMAGE DESCRIPTION (if available):\par
\texttt{\{IMAGE\_DESCRIPTION\}}\par

VISUAL STATEMENTS (from the model's reasoning):\par
1.~\texttt{\{STATEMENT\_1\}}\par
2.~\texttt{\{STATEMENT\_2\}}\par
3.~\texttt{\{STATEMENT\_3\}}\par
...\par

\end{tcolorbox}
\end{figure*}

\begin{figure*}[t]
\centering
\begin{tcolorbox}[
    width=\textwidth,
    colback=gray!10,
    colframe=gray,
    coltitle=white,
    fonttitle=\bfseries,
    title=Reasoning System Prompt (Ablation Study),
    attach boxed title to top left,
    boxed title style={
        colback=gray!50!black,
        boxrule=0pt,
        arc=0pt,
        outer arc=0pt,
        top=3pt,
        bottom=3pt,
        left=4pt,
        right=4pt
    },
    enhanced,
    sharp corners,
    left=6pt,
    right=6pt,
    top=6pt,
    bottom=6pt,
    halign title=flush left,
    before upper=\raggedright
]
\small

A conversation between a User and an Assistant.  
The User asks a question about an image, and the Assistant solves it.  
The Assistant first thinks through the reasoning process internally and then provides the final answer.

The reasoning process and the final answer must be enclosed within  
\texttt{<think>} \texttt{</think>} and \texttt{<answer>} \texttt{</answer>} tags, respectively.  
For example:\par
\texttt{<think>} reasoning process here \texttt{</think>}  
\texttt{<answer>} answer here \texttt{</answer>} \par

You are given:\par
- An image,\par
- A question about the image.\par

Your task:\par
1. Carefully analyze the image.\par
2. Identify objects, attributes, and spatial relations relevant to the question.\par
3. Perform step-by-step visual reasoning.\par
4. Give a concise and correct final answer.\par

Follow these rules:\par
- The reasoning inside \texttt{<think>} should be detailed and explicit.\par
- The final answer inside \texttt{<answer>} should be short and definitive.\par
- Do not mention the image analysis process outside of the \texttt{<think>} block.\par

--------------------\par
QUESTION:\par
\texttt{\{QUESTION\_TEXT\}}\par

IMAGE:\par
\texttt{\{IMAGE\_PLACEHOLDER\}}\par

\end{tcolorbox}
\end{figure*}

\begin{figure*}[t]
\centering
\begin{tcolorbox}[
    width=\textwidth,
    colback=gray!10,
    colframe=gray,
    coltitle=white,
    fonttitle=\bfseries,
    title=Fine-Grained Reasoning Prompt (visualization),
    attach boxed title to top left,
    boxed title style={
        colback=gray!50!black,
        boxrule=0pt,
        arc=0pt,
        outer arc=0pt,
        top=3pt,
        bottom=3pt,
        left=4pt,
        right=4pt
    },
    enhanced,
    sharp corners,
    left=6pt,
    right=6pt,
    top=6pt,
    bottom=6pt,
    halign title=flush left,
    before upper=\raggedright
]
\small

You are an Assistant that performs \textbf{fine-grained visual reasoning}.  
You are given an image and a question. Your task is to examine the image carefully,
identify all visually relevant details, and use them to answer the question.

Your reasoning process must:\par
- Attend to \textbf{small visual cues} such as colors, textures, shapes, numbers, and spatial relations.\par
- Explicitly verify each visual observation before relying on it.\par
- Avoid hallucinating any object, attribute, or relation not present in the image.\par
- Connect visual details step by step to form a coherent reasoning chain.\par

The reasoning process must be written inside \texttt{<think>} \texttt{</think>}.\par
The final short answer must be written inside \texttt{<answer>} \texttt{</answer>}.\par

Example format:\par
\texttt{<think>} detailed visual reasoning here \texttt{</think>}\par
\texttt{<answer>} final answer here \texttt{</answer>}\par

Instructions:\par
1. Begin by describing all key visual details relevant to the question.\par
2. Compare objects, count elements, check colors, sizes, and spatial layouts.\par
3. Use concrete grounding from the image before making any inference.\par
4. Produce the final answer only after completing the visual reasoning.\par

--------------------\par
QUESTION:\par
\texttt{\{QUESTION\_TEXT\}}\par

IMAGE:\par
\texttt{\{IMAGE\_PLACEHOLDER\}}\par

\end{tcolorbox}
\end{figure*}

\twocolumn

\clearpage

{
    \small
    \bibliographystyle{ieeenat_fullname}
    \bibliography{ref}
}
